\documentclass[pdflatex,sn-mathphys-num]{sn-jnl}

\usepackage{hhline}
\usepackage{makecell}
\usepackage{graphicx}%
\usepackage{multirow}%
\usepackage{amsmath,amssymb,amsfonts}%
\usepackage{amsthm}%
\usepackage{soul}
\usepackage{mathrsfs}%
\usepackage[title]{appendix}%
\usepackage{textcomp}%
\usepackage{manyfoot}%
\usepackage{booktabs}%
\usepackage{algorithm}%
\usepackage{algorithmicx}%
\usepackage{algpseudocode}%
\usepackage{listings}%
\usepackage{array}    
\usepackage{pifont}
\usepackage{subcaption}
\usepackage{ragged2e}
\usepackage[table]{xcolor}
\usepackage{enumitem}
\usepackage[most]{tcolorbox}
\usepackage{tabularx}

\definecolor{darkgreen}{RGB}{0,150,0}
\definecolor{darkred}{RGB}{200,0,0} 
\definecolor{ExpertColor}{HTML}{9370DB}
\definecolor{LLMColor}{HTML}{87CEEB}
\definecolor{CrowdColor}{HTML}{FF6961}

\usepackage{tikz}
\usepackage{caption}
\newcolumntype{L}[1]{>{\RaggedRight\arraybackslash}p{#1}}
\usepackage[export]{adjustbox} 
\usepackage[most]{tcolorbox} 
\usepackage{graphicx}      
\usepackage{tabularx}      
\usepackage{lipsum}  
\usepackage{relsize}
\usepackage{fontawesome5}  
\definecolor{iosblue}{RGB}{0, 122, 255}
\definecolor{iosgray}{RGB}{229, 229, 234}
\definecolor{placeholdergray}{RGB}{209, 213, 219}
\definecolor{placeholdericon}{RGB}{107, 114, 128}

\theoremstyle{thmstyleone}%
\theoremstyle{thmstyletwo}%

\theoremstyle{thmstylethree}%

\usepackage{etoolbox}

\begin{document}

\title[When LLMs are Reliable for Empathic Communication]{When Large Language Models are Reliable for Judging Empathic Communication}

\author*[1,2]{\fnm{Aakriti} \sur{Kumar}}\email{aakriti.kumar@kellogg.northwestern.edu}

\author[1]{\fnm{Nalin} \sur{Poungpeth}}

\author[3]{\fnm{Diyi} \sur{Yang}}

\author[4]{\fnm{Erina} \sur{Farrell}}

\author[5]{\fnm{Bruce} \sur{Lambert}}

\author*[1,2,6]{\fnm{Matthew} \sur{Groh}}\email{matthew.groh@kellogg.northwestern.edu}

\affil[1]{\orgdiv{Kellogg School of Management}, \orgname{Northwestern University}}

\affil[2]{\orgdiv{Northwestern Institute for Complex Systems}, \orgname{Northwestern University}}

\affil[3]{\orgdiv{Department of Computer Science}, \orgname{Stanford University}}

\affil[4]{\orgdiv{Department of Communication Arts and Sciences}, \orgname{Pennsylvania State University}}

\affil[5]{\orgdiv{Department of Communication Studies}, \orgname{Northwestern University}}

\affil[6]{\orgdiv{Department of Computer Science}, \orgname{Northwestern University}}

\abstract{

Large language models (LLMs) excel at generating empathic responses in text-based conversations. But, how reliably do they judge the nuances of empathic communication? We investigate this question by comparing how experts, crowdworkers, and LLMs annotate empathic communication across four evaluative frameworks drawn from psychology, natural language processing, and communications applied to 200 real-world conversations where one speaker shares a personal problem and the other offers support. Drawing on 3,150 expert annotations, 2,844 crowd annotations, and 3,150 LLM annotations, we assess inter-rater reliability between these three annotator groups. We find that expert agreement is high but varies across the frameworks' sub-components depending on their clarity, complexity, and subjectivity. We show that expert agreement offers a more informative benchmark for contextualizing LLM performance than standard classification metrics. Across all four frameworks, LLMs consistently approach this expert level benchmark and exceed the reliability of crowdworkers. These results demonstrate how LLMs, when validated on specific tasks with appropriate benchmarks, can support transparency and oversight in emotionally sensitive applications including their use as conversational companions.}


\renewcommand{\thesubfigure}{\textbf{\Alph{subfigure}}}
\captionsetup[sub]{labelformat=simple}  
\newcommand{\mybold}[1]{\textbf{#1}}
\newcommand{\highlightcell}[2]{\makebox[90\%][c]{\cellcolor{#1}#2}}

\maketitle


Based on how people interact with large language models (LLMs), LLMs appear to show signs of emotional intelligence in their skillful recognition of people's emotions and empathic responses to them~\cite{salovey1990emotional, picard2000affective}. In particular, there is growing evidence that LLMs can generate responses that people perceive as empathic, often rating them higher than human responses~\cite{sorin2024large, inzlicht2024praise}. People report feeling heard and understood by LLM responses~\cite{yin2024ai,lee2024large,rubin2024value}, rate LLM responses as higher than humans' in cognitive re-appraisal~\cite{li2024skill}, compassion~\cite{ovsyannikova2025third}, and empathy ~\cite{sharma2023human, ayers2023comparing}, identify LLM responses as helping them to reduce negative thinking~\cite{swain2025}, and experience reduced anxiety and depression after interacting with an LLM designed for therapy~\cite{heinz2025randomized}. This machine capacity for empathic communication has sparked applications in contexts ranging from customer service~\cite{hayawi2024care, hara2024autonomous} to health care~\cite{steenstra2025scaffolding}. However, the ability to generate empathic responses is distinct from the ability to evaluate them, and prior work suggests that the generative and evaluative capabilities of LLMs may not always align~\cite{west2023generative,liu2023benchmarking, dell2023navigating}. Deploying these models with accountability and transparency therefore requires addressing an open question: How reliably can machines judge the nuances of empathic communication? 

Misjudging an LLM's empathic communication skills can have serious consequences ranging from ineffective support, distrust, and bias to significant unintended harm. For example, LLMs-as-companions sometimes offer unrealistic, exaggerated responses to low stakes emotional situations~\cite{roshanaei2024talk}. Given that LLMs respond differently to different inputs, researchers have shown that levels of empathic support systematically vary across demographic groups~\cite{gabriel2024can}. Moreover, significant harm can occur in chats with LLMs-as-companions, encouraging delusional thinking~\cite{moore2025expressing}, increasing emotional dependence on the LLMs~\cite{fang2025ai}, and even encouraging people to commit suicide ~\cite{turkle2024chatbot, xiang2023ai, adam2025supportive}. These incidents, along with the possibility that LLMs' empathic communication skills unexpectedly change  due to emergent misalignment~\cite{betley2025emergent}, highlight the urgent need for rigorous and reliable evaluation before and during deployment of AI systems in sensitive user-facing contexts. By systematically examining LLM performance in judging empathic communication relative to expert and novice humans, we demonstrate LLMs' capacity for identifying aligned and misaligned responses, thereby scoping the potential for automated assessments of empathic communication. 

We focus on evaluating \textbf{empathic communication}\footnote{Empathic communication (also referred to comforting, emotional support, empathic listening when offering support, and perceived empathy when receiving support) refers to the act of acknowledging and responding sensitively to a person's emotions, experiences, and perspectives with the goal to make them feel understood~\cite{suchman1997model, burleson2003emotional, covey20207, nambisan2011information, groh2022computational}} in text-based conversations. We intentionally do not consider empathy-as-a-trait, which is often defined as how well an individual can understand the emotions and experiences of another~\cite{mehrabian1972measure, jordan2016empathy, konrath2018development} because empathy-as-a-trait in LLMs can lead to a paradox of semantics~\cite{shteynberg2024does, perry2023ai}. Instead, we consider how words may make someone feel empathized with based on frameworks designed to evaluate empathic communication. In particular, we examine a set of three frameworks associated with peer reviewed papers and publicly available datasets – Empathetic Dialogues~\cite{rashkin2018towards}, EPITOME~\cite{sharma2020computational}, and Perceived Empathy~\cite{yin2024ai} – that have been highly cited in natural language processing (NLP) and psychology. In addition, we introduce a fourth framework (accompanied by the Lend an Ear pilot dataset) grounded in motivational interviewing~\cite{moyers2014motivational}, empathic listening \cite{covey20207, drollinger2006development}, physician-patient empathic communication~\cite{suchman1997model, mercer2004consultation, bylund2005examining}, relationship scoring~\cite{barrett2015relationship}, and empathy trainings~\cite{teding2016efficacy}. Each framework identifies different questions for annotating sub-components of empathic communication or lack thereof. While there is no universally agreed upon framework for empathic communication \cite{suchman1997model, bylund2005examining}, these four evaluative frameworks provide a starting point for explicitly evaluating sub-components of empathic communication in text based conversations. 

\begin{figure*}[htbp]

    \centering

    \begin{subfigure}{\textwidth}
        \centering
        \adjustbox{max width=\linewidth, valign=t}{%
            \includegraphics[width=\linewidth]{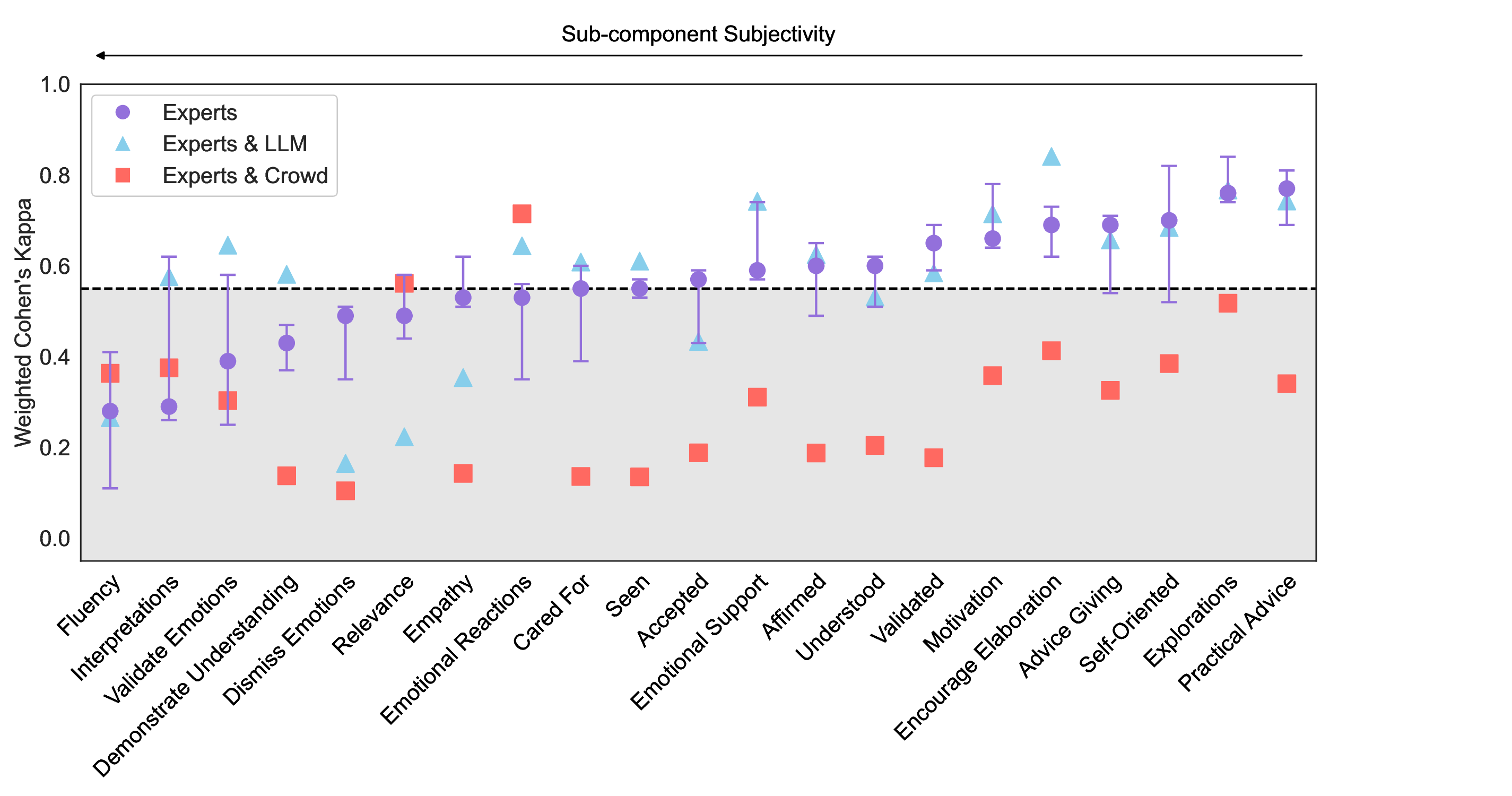}%
        }
        \label{subfig:kappa_plot}
    \end{subfigure}

    \vspace{.5em}
    \begin{subfigure}{.9\textwidth}
        \centering
        \newcolumntype{R}[1]{>{\raggedleft\arraybackslash}p{#1}}
        \newcolumntype{L}[1]{>{\raggedright\arraybackslash}p{#1}}
        \adjustbox{max width=\linewidth, valign=c}{%
        \resizebox{\textwidth}{!}{%
        \begin{tabular}{L{4cm} >{\raggedright\arraybackslash\scriptsize}p{15.5cm}}
        \toprule
        \textbf{Sub-component} & \normalsize\textbf{Annotation Question} \\ 
        \midrule
        \multicolumn{2}{l}{\textbf{Empathetic Dialogues} (5 point Likert scale)} \\  
        \midrule
        Empathy & Did the responses demonstrate an understanding of the feelings of the person sharing their experience? \\
        \arrayrulecolor{black!30}\midrule
        Relevance & Did the responses seem appropriate to the conversation? Were they on-topic? \\
        \arrayrulecolor{black!30}\midrule
        Fluency & Were the responses easy to understand? Did the language seem accurate? \\
        \arrayrulecolor{black}\midrule
        \multicolumn{2}{l}{\textbf{EPITOME} (3 point scale)} \\
        \arrayrulecolor{black}\midrule
        Emotional Reactions & Does the response express or allude to warmth, compassion, concern, or similar feelings of the responder towards the seeker? \\
        \arrayrulecolor{black!30}\midrule
        Explorations & Does the response make an attempt to explore the seeker's experiences and feelings? \\
        \arrayrulecolor{black!30}\midrule
        Interpretations & Does the response communicate an understanding of the seeker's experiences and feelings? \\
        \arrayrulecolor{black}\midrule
        \multicolumn{2}{l}{\textbf{Perceived Empathy} (7 point Likert scale)} \\
        \arrayrulecolor{black}\midrule
        Understood & To what extent do you think reading the response would make the discloser feel understood? \\
        \arrayrulecolor{black!30}\midrule
        Validated & To what extent do you think reading the response would make the discloser feel validated? \\
        \arrayrulecolor{black!30}\midrule
         
        Affirmed & To what extent do you think reading the response would make the discloser feel affirmed? \\
        \arrayrulecolor{black!30}\midrule
         
        Accepted & To what extent do you think reading the response would make the discloser feel accepted? \\
        \arrayrulecolor{black!30}\midrule
         
        Cared For & To what extent do you think reading the response would make the discloser feel cared for? \\
        \arrayrulecolor{black!30}\midrule
         
        Seen & To what extent do you think reading the response would make the discloser feel seen? \\
        \arrayrulecolor{black!30}\midrule
         
        Emotional Support & To what extent the responder provided emotional support (e.g., offers of reassurance, expressions of concern)? \\
        \arrayrulecolor{black!30}\midrule
         
        Practical Advice & To what extent the responder provided practical support (e.g., advice, suggestions of courses of action, offers of direct assistance)? \\
        \arrayrulecolor{black!30}\midrule
         
        Motivation & Rate how much effort the responder put into writing the response. \\
        \arrayrulecolor{black}\midrule
        \multicolumn{2}{l}{\textbf{Lend an Ear} (5 point Likert scale)} \\
        \arrayrulecolor{black}\midrule
        Validated Emotions & To what extent did the supporter validate their partner's emotions? \\
        \arrayrulecolor{black!30}\midrule
         
        Encouraging Elaboration & To what extent did the supporter ask questions and encourage their partner to elaborate on their experiences and emotions? \\
        \arrayrulecolor{black!30}\midrule
         
        Demonstrating Understanding & To what extent did the supporter use paraphrasing to demonstrate their understanding of what their partner is going through? \\
        \arrayrulecolor{black!30}\midrule
         
        Advice Giving & To what extent did the supporter provide unsolicited advice to their partner. \\
        \arrayrulecolor{black!30}\midrule
         
        Self-Oriented & To what extent did the supporter shift the focus to themselves? \\
        \arrayrulecolor{black!30}\midrule
         
        Dismissing Emotions & To what extent did the supporter dismiss their partner's emotions? \\
        \arrayrulecolor{black}\bottomrule
        \end{tabular}}}
    \end{subfigure}

    \caption{\textbf{Reliability Across Annotator Pairs and Sub-Components.} \textbf{Top}: Inter-rater reliability (quadratically weighted $\kappa_w$) across annotator pairs for each empathic communication sub-components. In the evaluation of experts with each other, the circle represents the median $\kappa_w$ and the error bar represents the $\kappa_w$ range between the three pairs. Experts comparisons with the LLM (Gemini 2.5 Pro) and crowd compare median expert annotations with the LLM and crowd, respectively. \textbf{Bottom}: Empathic communication frameworks with verbatim annotation questions.}
    \label{fig:stacked_overview}

\end{figure*}

Each of the four datasets examined in this paper is made up of dyadic conversations in which one partner shares a personal difficulty and another attempts to responds supportively. However, the context of conversations in each dataset varies across a number of dimensions such as the number of conversational turns, whether the interaction was synchronous or asynchronous, the topic of the conversation and how it was chosen, and how participants were recruited (see Table~\ref{tab:sub_context_comparison} and Methods for details). Most importantly for our evaluation, these frameworks for annotating empathic communication differ in the number and kind of questions asked (see Figure~\ref{fig:stacked_overview} for a verbatim list of questions included in each framework). The NLP-based frameworks, Empathetic Dialogues and EPITOME, each include three questions. However, Empathetic Dialogues has a single question directly related to empathic communication whereas EPITOME's three questions are grounded in Motivational Interviewing Skill Code (MISC)~\cite{de2005motivational}. The Perceived Empathy framework draws on the psychology of shared realities~\cite{hardin1996shared,elnakouri2023together} and offers the highest level of granularity with nine overlapping dimensions of evaluation. Finally, the Lend an Ear framework includes six questions that address prescriptive and proscriptive guidelines derived from normative models of empathic communication. For all four of these frameworks and associated datasets, researchers recruited crowdworkers to annotate conversations.

Crowdsourced annotations for empathic communication offer a useful starting point for evaluating large conversational datasets but impose significant limitations. First, disagreement is common among crowdworkers. Addressing disagreement with detailed guidelines does not necessarily increase reliability because crowdworkers annotating microtasks tend not to read complex annotation guidelines~\cite{aroyo2015truth}. Second, there is evidence that lay people are generally unaware of the nuances of empathic communication. When research participants are asked to generate comforting messages, the messages that the majority of participants express only moderate rather than high empathy accordingly to a theoretically-grounded rubric for evaluating comforting~\cite{macgeorge2003skill, macgeorge2017influence, samter2016coding}. To the extent that the typical support provider is unable (or unwilling) to produce highly empathic messages, they may also fail to recognize or appreciate the value of those messages when others generate them. One approach to addressing skill issues arising with crowdworkers is to recruit annotators with communications expertise. While reasonable experts will not always agree on the exact annotation for any particular coding scheme, research on another framework for empathic communication, Verbal Person Centeredness~\cite{macgeorge2017influence}, shows that trained annotators can achieve a Kripendorff's $\alpha$ of 0.73 when evaluating 15 minute conversations, which generally matches the inter-rater reliability of experts annotating briefer written support messages~\cite{macgeorge2003skill, macgeorge2012predicting}.

LLMs-as-judge represents an alternative annotation approach that has the potential to address both problems of quality in crowdsourcing and scalability in expert annotations.  In many cases, when prompted with expert-designed annotation guidelines, LLMs demonstrate strong alignment with human evaluators~\cite{ziems2024can, movva2024annotation}. LLMs have also been effective at annotating fundamental elements of therapy and counseling. They have been used to identify psychotherapy techniques from therapist and client utterances in therapy sessions~\cite{chiu2024computational}, annotate conversational markers in online text-based counseling that correlate with effective therapeutic engagement~\cite{li2024understanding}, and identify storytelling elements that contribute to empathic engagement in personal narratives~\cite{shen2024heart}. However, LLM-based annotations are prone to variability across runs~\cite{barrie2024replication}, inconsistencies across contexts~\cite{gu2024survey}, apparently sensible yet wrong answers~\cite{zhou2024larger}, miscalibrated confidence~\cite{Steyvers2025}, and a series of human-like cognitive biases~\cite{ye2024justice, balog2025rankers}. 
 
Given these mixed prior findings, LLM's reliability specifically for annotating empathic communication compared to human (expert and crowd) benchmarks remains largely unexplored and calls for systematic investigation. Our research addresses this gap, motivated by the need for a reliable and scalable method for creating accountability and transparency into how LLMs communicate with humans. Specifically, we investigate how reliably experts, crowds, and LLMs evaluate empathic communication in text-based conversations across four different frameworks and conversational contexts. By collecting 1050 annotations each from three different experts, a variable numbers of crowdworkers, and three different state-of-the-art LLMs, we contextualize LLMs' performance relative to experts and crowds across multiple contexts. This rigorous evaluation of LLMs-as-judge against inter-rater reliability of experts and crowds reveals blind spots that would otherwise emerge when applying classification metrics to assumed ground truth annotations. Finally, we highlight the reliability (or lack thereof) of each sub-component of each framework and offer a qualitative analysis that reveals where disagreements tends to arise.

\begin{table}[htbp]
    \centering
    \resizebox{\linewidth}{!}{
    {\fontsize{15}{14}\selectfont    
    \begin{tabular}{@{}lcccc@{}}
    \toprule
     & \textbf{Empathetic Dialogues} & \textbf{EPITOME} & \textbf{Perceived Empathy} & \textbf{Lend an Ear} \\
    \midrule
    \textbf{Contextual Themes} & & & & \\
    \midrule
    Multi-Turn Conversation  & \textcolor{darkgreen}{\ding{51}} & \textcolor{darkred}{\ding{55}} & \textcolor{darkred}{\ding{55}} & \textcolor{darkgreen}{\ding{51}} \\
    Synchronous Conversation  & \textcolor{darkred}{\ding{55}} & \textcolor{darkred}{\ding{55}} & \textcolor{darkred}{\ding{55}} &  \textcolor{darkgreen}{\ding{51}} \\
    Predetermined Topic  & \textcolor{darkred}{\ding{55}} & \textcolor{darkred}{\ding{55}} &     \textcolor{darkred}{\ding{55}} & \textcolor{darkgreen}{\ding{51}} \\
    Human Sharing Trouble   & \textcolor{darkgreen}{\ding{51}} & \textcolor{darkgreen}{\ding{51}} & \textcolor{darkgreen}{\ding{51}}  &  \textcolor{darkred}{\ding{55}} \\
    Human Providing Support  & \textcolor{darkgreen}{\ding{51}} & \textcolor{darkgreen}{\ding{51}} & \textcolor{darkgreen}{\ding{51}}  & \textcolor{darkgreen}{\ding{51}}\\
    \midrule
    \textbf{Conversational Statistics} & & & &  \\
    \midrule
    Conversational Turns (Median)  & 5 & 2 & 2 & 13 \\
    Words per Conversation (Median)  & 75 & 72 & 226 & 385 \\
    \midrule
    \textbf{Dataset Information} & & & &  \\
    \midrule
    Date Published & 2018 & 2020 & 2024 & 2025 \\
    Number of Conversations & 24,850 & 3,081 & 501 & 50 \\
    Availability  & Public & Public & On request & Public \\
    Participants & MTurk & Reddit Users & Prolific & Prolific \\
    Platform  & ParlAI & Reddit & Qualtrics & Lend an Ear \\
    \bottomrule
    \end{tabular}}}
    \caption{\textbf{Conversational Contexts across Datasets.} Key information on all available datasets. EPITOME also includes a second dataset from TalkLife that is not included here because access is only available upon request to the TalkLife company~\cite{sharma2020computational} The full Perceived Empathy dataset is available upon request to the authors of the associated paper~\cite{yin2024ai}.}
    \label{tab:sub_context_comparison}

\end{table}

\section*{Results}\label{results}

\begin{figure}[htb]
    \centering
    \captionsetup{justification=raggedright, singlelinecheck=false, skip=2pt, font=small}
    \includegraphics[width=\linewidth]{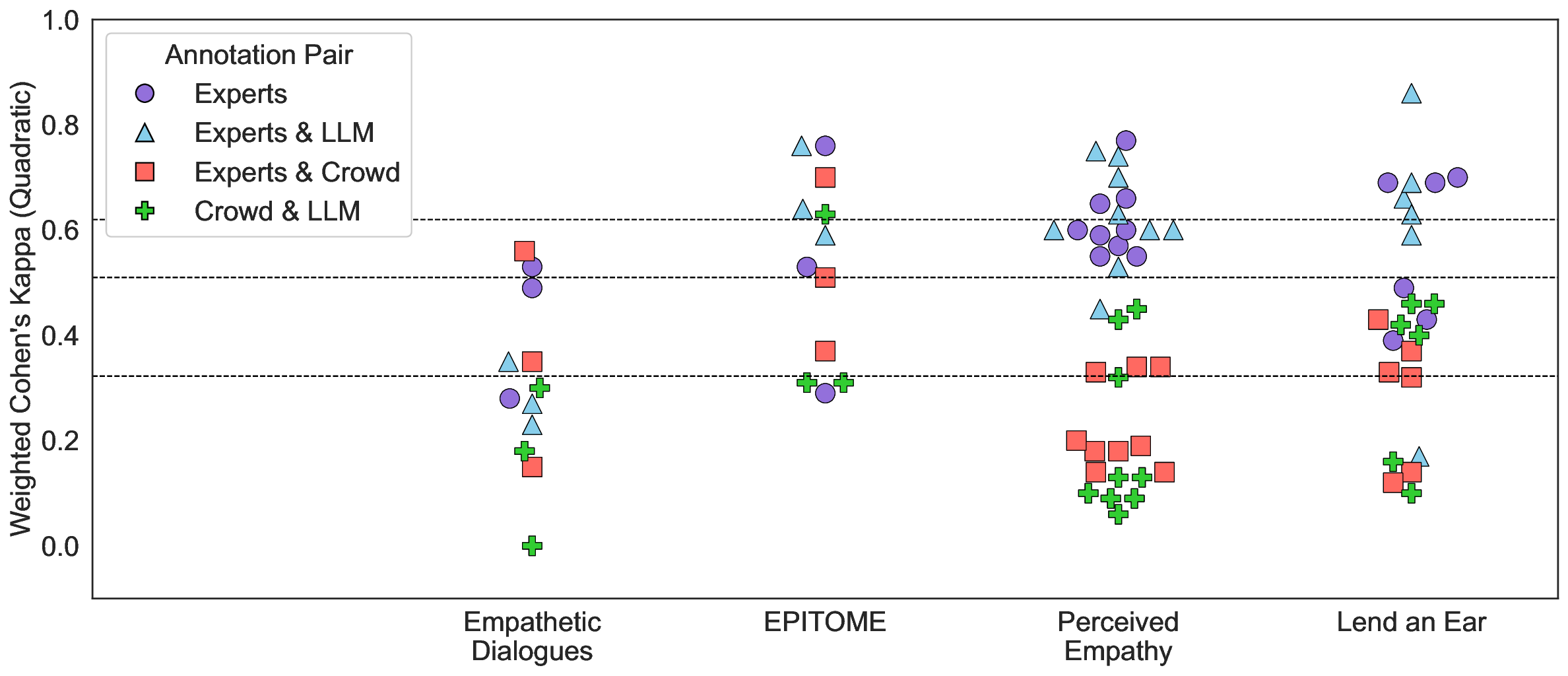}
    \caption{\mybold{Reliability across Annotator Pairs and Frameworks.} Inter-rater reliability (quadratically weighted $\kappa_w$)) for empathic communication across annotator pairs for each sub-component grouped by framework. In the evaluation of experts with each other,  circle represents the the median $\kappa_w$ between the three expert pairs. Experts comparisons with the LLM (Gemini 2.5 Pro) and crowd compare median expert annotations with the LLM and crowd, respectively. The beeswarm plot ensures all data points are visible by applying a horizontal jitter to avoid overlap. Dotted lines represent the 25th, 50th, and 75th percentiles of Weighted Cohen's Kappa. 
    }
    \label{fig:bee_kappa}
\end{figure}

\subsection*{Expert annotations offer a benchmark for reliability}
Experts show relatively high inter-rater reliability across sub-components (Figure \ref{fig:bee_kappa}) with Krippendorff's alpha between experts ranging from $0.29$ to $0.78$ (median = $0.55$). Similarly, Cohen's weighted kappa ($\kappa_w$) between experts ranges from $0.11$ to $0.84$ (median $= 0.58$), with most values between $0.49$ and $0.69$ (IQR). Following traditional interpretations~\cite{landis1977measurement}, this corresponds to substantial agreement ($\kappa_w\geq 0.60$) for 9 sub-components, moderate agreement ($0.40 \leq \kappa_w \leq 0.60$) for 6 sub-components, and fair agreement ($0.20 \leq \kappa_w \leq 0.40$) for 6 sub-components. Additional details, including pairwise expert reliability and Krippendorff’s alpha across sub-components, can be found in Appendix Figures~\ref{fig:expertIRR} and \ref{fig:k-alpha}.

Given the inherent arbitrariness in mapping Cohen’s Kappa values into agreement strengths, we follow the approach of Schober et al. (2018)~\cite{schober2018correlation}, and ground our interpretation of inter-rater reliability within the specific context of evaluating empathic communication. To provide a meaningful threshold, we use a practical benchmark for high agreement using the median expert Kappa ($\kappa_w = 0.58$) across sub-components. This median value reflects substantial agreement among experts and is used as a reference threshold in Figure \ref{fig:stacked_overview} (dashed line).

Furthermore, we use the observed expert reliability for each sub-component as a benchmark for interpreting reliability among other annotator pairs. This allows us to interpret Kappa values for other annotator pairs (e.g., expert-LLM, expert-crowd) relative to the agreement achieved by experts in evaluating empathic communication.

\subsection*{LLMs demonstrate near-expert reliability}
When guided by a prompt that combines a detailed framework of empathic communication with three examples of expert annotations, LLMs annotations show the highest reliability with expert annotations. Appendix Table~\ref{ED:prompt-variants} shows that this combination prompt led to higher reliability with expert annotations compared to zero-shot and framework-only variants of the prompt. We use this prompt throughout our analyses. Across three different LLMs, Gemini 2.5 Pro (gemini-2.5-pro-preview-03-25), ChatGPT 4o (gpt-4o-2024-08-06), and Claude 3.7 Sonnet (claude-3-7-sonnet-20250219), we observed similar annotation performance with Krippendorff's alpha ranging from $0.51$ to $0.75$ (median = $0.60$). Appendix Figure~\ref{fig:llm-kappa} illustrates LLM Krippendorff's alpha across sub-components and Appendix Table~\ref{ED:llms-kappa-median} details LLM reliability with expert annotations using few-shot and Chain-of-Thought (CoT) prompting methods. Given the relatively high inter-rater reliability between LLMs, all subsequent analyses in this paper use annotations from Gemini 2.5 Pro based on the few-shot prompt shared in the Appendix. 

Compared to our expert agreement benchmark, reliability between experts and LLMs closely followed expert reliability with $\kappa_w$ between experts and LLMs ranging from $0.17$ to $0.86$ (median $= 0.60$), with most values between $0.49$ and $0.70$ (IQR). Expert-LLM pairs exceeded the high-agreement threshold in 15 of 21 sub-components (70\%), particularly in all sub-components where experts showed high agreement (Figure \ref{fig:stacked_overview}). 

Additionally, patterns of variability in LLM annotations closely mirrored expert variability across sub-components. Specifically, we find a Pearson correlation of $0.67$ between expert-expert and expert-LLM kappa scores. In contrast, the correlation between expert-expert and expert-crowd kappa scores was $0.17$, indicating higher discrepancies between experts and crowdworkers. Appendix includes a robustness analysis of LLM annotations that compares LLM performance across repeated runs, conversational contexts, and verbosity of conversation turns.

\subsection*{Reliability varies across frameworks}
We find that inter-rater reliability varies widely across frameworks. The mean expert $\kappa_w$ for Empathetic Dialogues, EPITOME, Perceived Empathy, and Lend an Ear pilot are $0.41$, $0.46$, $0.60$, $0.55$, respectively. This variability suggests that qualitative differences across frameworks, such as context-dependence and operational clarity of sub-components influence the reliability of experts, crowdworkers, and LLMs alike.

For the Empathetic Dialogues and Perceived Empathy frameworks, we observe high multicollinearity among expert annotations, assessed via variance inflation factors (VIF). Specifically, we find VIF values greater than 10 in all sub-components of Empathetic Dialogues, and 8 out of 9 sub-components in Perceived Empathy. This suggests substantial overlap in how each annotator interpreted and rated these conceptually related dimensions, and calls into question the marginal added value of these frameworks' sub-components (see Appendix Figure~\ref{fig:vif} for detailed VIF results).

Furthermore, the Empathetic Dialogues framework showed the lowest expert reliability, indicating ambiguity or subjectivity in its operationalizations. Similarly, EPITOME, despite being widely used \cite{sharma2021towards, lee2022does, sharma2023human, gabriel2024can}, achieves relatively low expert reliability on two of its three sub-components due to insufficient clarity and operational definitions (see Table \ref{tab:kappa_table}).

\subsection*{Reliability varies across sub-components}
Expert annotations showed relatively high agreement on several sub-components, primarily in the Perceived Empathy and Lend an Ear pilot datasets. These include the ``Self-Oriented'', ``Advice-Giving'', ``Encouraging Elaboration'' from the Lend an Ear experiment, ``Practical Advice'', ``Emotional Support'', ``Understood'', ``Validated'' and ``Motivation'' from the Perceived Empathy dataset, and ``Explorations'' from the EPITOME dataset. Detailed Cohen's Kappa values for each annotator pair and sub-component are provided in Table \ref{tab:kappa_table}.

We find that expert agreement was generally higher for sub-components characterized by clear linguistic or behavioral markers. For instance, involving identifiable linguistic markers such as identifying questions designed to encourage elaboration (``Encouraging Elaboration'' in Lend an Ear, median $\kappa_w=0.69$) and exploring the support seeker's context (``Explorations''  in EPITOME, median $\kappa_w=0.76$) were annotated with high reliability. Additionally, components focused on behavioral markers such as advice-giving, namely ``Practical Advice'' in Perceived Empathy (median $\kappa_w=0.77$) and ``Advice-Giving'' in Lend an Ear (median $\kappa_w=0.69$) were also relatively straightforward to annotate for experts. 

Conversely, sub-components requiring more subjective judgment about both the speaker’s intentions and the listener’s emotional state were prone to lower inter-rater reliability even for experts. Note that, intentions are inherently unobservable in conversation transcripts, any sub-component that requires annotators to judge them rather than the contents of message increases the risk of subjective guesswork. For instance, low expert agreement was observed for ``Interpretations'' (median $\kappa_w=0.29$) and ``Demonstrating Understanding'' in Lend an Ear (median $\kappa_w=0.43$). Our findings thus highlight how operational clarity is key to consistent annotation, especially for less explicit behaviors.

\begin{table}[h!]
\centering
\resizebox{\textwidth}{!}{
\begin{tabular}{l|c|c|c|c|c|c|}
\toprule
\textbf{Sub-component} & \makecell{\textbf{Expert1 \&} \\ \textbf{Expert2}} & \makecell{\textbf{Expert2 \&} \\ \textbf{Expert3}} & \makecell{\textbf{Expert3 \&} \\ \textbf{Expert1}} & \makecell{\textbf{Experts \&} \\ \textbf{Crowd}} & \makecell{\textbf{Experts \&} \\ \textbf{LLM}} & \makecell{\textbf{Crowd \&} \\ \textbf{LLM}} \\\midrule
\multicolumn{5}{l}{\textbf{Empathetic Dialogues}} \\
\cmidrule{1-7}
Empathy & 0.51\textsuperscript{\relsize{-3}***} & \cellcolor{green!25}\textbf{0.62\textsuperscript{\relsize{-3}***}} & 0.53\textsuperscript{\relsize{-3}***} & 0.15\textsuperscript{\relsize{-3}*} & 0.35\textsuperscript{\relsize{-3}***} & 0.00 \\
Fluency & 0.28 & 0.11 & 0.41\textsuperscript{\relsize{-3}***} & 0.35\textsuperscript{\relsize{-3}*} & 0.27 & 0.30 \\
Relevance & 0.44\textsuperscript{\relsize{-3}*} & \cellcolor{green!25}\textbf{0.58\textsuperscript{\relsize{-3}**}} & 0.49\textsuperscript{\relsize{-3}**} & 0.56\textsuperscript{\relsize{-3}***} & 0.23 & 0.18 \\
\midrule
\multicolumn{5}{l}{\textbf{EPITOME}} \\
\cmidrule{1-7}
Emotional Reactions & 0.35\textsuperscript{\relsize{-3}**} & 0.56\textsuperscript{\relsize{-3}***} & 0.53\textsuperscript{\relsize{-3}***} & \cellcolor{green!25}\textbf{0.70\textsuperscript{\relsize{-3}***}} & \cellcolor{green!25}\textbf{0.64\textsuperscript{\relsize{-3}***}} & \cellcolor{green!25}\textbf{0.63\textsuperscript{\relsize{-3}***}} \\
Explorations & \cellcolor{green!25}\textbf{0.76\textsuperscript{\relsize{-3}***}} & \cellcolor{green!25}\textbf{0.84\textsuperscript{\relsize{-3}***}} & \cellcolor{green!25}\textbf{0.74\textsuperscript{\relsize{-3}***}} & 0.51\textsuperscript{\relsize{-3}***} & \cellcolor{green!25}\textbf{0.76\textsuperscript{\relsize{-3}***}} & 0.31\textsuperscript{\relsize{-3}**} \\
Interpretations & 0.26\textsuperscript{\relsize{-3}**} & \cellcolor{green!25}\textbf{0.62\textsuperscript{\relsize{-3}***}} & 0.29\textsuperscript{\relsize{-3}***} & 0.37\textsuperscript{\relsize{-3}**} & \cellcolor{green!25}\textbf{0.59\textsuperscript{\relsize{-3}***}} & 0.31\textsuperscript{\relsize{-3}**} \\
\midrule
\multicolumn{5}{l}{\textbf{Perceived Empathy}} \\
\cmidrule{1-7}
Understood & \cellcolor{green!25}\textbf{0.60\textsuperscript{\relsize{-3}***}} & \cellcolor{green!25}\textbf{0.62\textsuperscript{\relsize{-3}***}} & 0.51\textsuperscript{\relsize{-3}***} & 0.20\textsuperscript{\relsize{-3}*} & 0.53\textsuperscript{\relsize{-3}***} & 0.06 \\
Validated & \cellcolor{green!25}\textbf{0.59\textsuperscript{\relsize{-3}***}} & \cellcolor{green!25}\textbf{0.69\textsuperscript{\relsize{-3}***}} & \cellcolor{green!25}\textbf{0.65\textsuperscript{\relsize{-3}***}} & 0.19\textsuperscript{\relsize{-3}*} & \cellcolor{green!25}\textbf{0.60\textsuperscript{\relsize{-3}***}} & 0.13 \\
Affirmed & 0.49\textsuperscript{\relsize{-3}***} & \cellcolor{green!25}\textbf{0.65\textsuperscript{\relsize{-3}***}} & \cellcolor{green!25}\textbf{0.60\textsuperscript{\relsize{-3}***}} & 0.18\textsuperscript{\relsize{-3}*} & \cellcolor{green!25}\textbf{0.63\textsuperscript{\relsize{-3}***}} & 0.13 \\
Accepted & 0.43\textsuperscript{\relsize{-3}***} & \cellcolor{green!25}\textbf{0.58\textsuperscript{\relsize{-3}***}} & 0.57\textsuperscript{\relsize{-3}***} & 0.18\textsuperscript{\relsize{-3}**} & 0.45\textsuperscript{\relsize{-3}***} & 0.09 \\
Cared For & 0.55\textsuperscript{\relsize{-3}***} & \cellcolor{green!25}\textbf{0.60\textsuperscript{\relsize{-3}***}} & 0.39\textsuperscript{\relsize{-3}***} & 0.14\textsuperscript{\relsize{-3}*} & \cellcolor{green!25}\textbf{0.60\textsuperscript{\relsize{-3}***}} & 0.10 \\
Seen & 0.55\textsuperscript{\relsize{-3}***} & 0.56\textsuperscript{\relsize{-3}***} & 0.54\textsuperscript{\relsize{-3}***} & 0.14 & \cellcolor{green!25}\textbf{0.60\textsuperscript{\relsize{-3}***}} & 0.09 \\
Emotional Support & \cellcolor{green!25}\textbf{0.74\textsuperscript{\relsize{-3}***}} & \cellcolor{green!25}\textbf{0.59\textsuperscript{\relsize{-3}***}} & \cellcolor{green!25}\textbf{0.59\textsuperscript{\relsize{-3}***}} & 0.33\textsuperscript{\relsize{-3}***} & \cellcolor{green!25}\textbf{0.75\textsuperscript{\relsize{-3}***}} & 0.43\textsuperscript{\relsize{-3}***} \\
Practical Advice & \cellcolor{green!25}\textbf{0.77\textsuperscript{\relsize{-3}***}} & \cellcolor{green!25}\textbf{0.69\textsuperscript{\relsize{-3}***}} & \cellcolor{green!25}\textbf{0.81\textsuperscript{\relsize{-3}***}} & 0.34\textsuperscript{\relsize{-3}***} & \cellcolor{green!25}\textbf{0.74\textsuperscript{\relsize{-3}***}} & 0.45\textsuperscript{\relsize{-3}***} \\
Motivation & \cellcolor{green!25}\textbf{0.66\textsuperscript{\relsize{-3}***}} & \cellcolor{green!25}\textbf{0.64\textsuperscript{\relsize{-3}***}} & \cellcolor{green!25}\textbf{0.78\textsuperscript{\relsize{-3}***}} & 0.34\textsuperscript{\relsize{-3}**} & \cellcolor{green!25}\textbf{0.70\textsuperscript{\relsize{-3}***}} & 0.32\textsuperscript{\relsize{-3}**} \\
\midrule
\multicolumn{5}{l}{\textbf{Lend an Ear}} \\
\cmidrule{1-7}
Validating Emotions & 0.25\textsuperscript{\relsize{-3}**} & \cellcolor{green!25}\textbf{0.58\textsuperscript{\relsize{-3}***}} & 0.39\textsuperscript{\relsize{-3}***} & 0.32\textsuperscript{\relsize{-3}**} & \cellcolor{green!25}\textbf{0.63\textsuperscript{\relsize{-3}***}} & 0.42\textsuperscript{\relsize{-3}***} \\
Demonstrating Understanding & 0.47\textsuperscript{\relsize{-3}***} & 0.43\textsuperscript{\relsize{-3}*} & 0.37 & 0.14\textsuperscript{\relsize{-3}*} & \cellcolor{green!25}\textbf{0.59\textsuperscript{\relsize{-3}***}} & 0.16\textsuperscript{\relsize{-3}**} \\
Encouraging Elaboration & \cellcolor{green!25}\textbf{0.69\textsuperscript{\relsize{-3}***}} & \cellcolor{green!25}\textbf{0.73\textsuperscript{\relsize{-3}***}} & \cellcolor{green!25}\textbf{0.62\textsuperscript{\relsize{-3}***}} & 0.43\textsuperscript{\relsize{-3}***} & \cellcolor{green!25}\textbf{0.86\textsuperscript{\relsize{-3}***}} & 0.40\textsuperscript{\relsize{-3}***} \\
Advice Giving & \cellcolor{green!25}\textbf{0.70\textsuperscript{\relsize{-3}***}} & \cellcolor{green!25}\textbf{0.69\textsuperscript{\relsize{-3}***}} & 0.54\textsuperscript{\relsize{-3}***} & 0.33\textsuperscript{\relsize{-3}**} & \cellcolor{green!25}\textbf{0.66\textsuperscript{\relsize{-3}***}} & 0.46\textsuperscript{\relsize{-3}***} \\
Self-Oriented & \cellcolor{green!25}\textbf{0.82\textsuperscript{\relsize{-3}***}} & 0.52\textsuperscript{\relsize{-3}***} & \cellcolor{green!25}\textbf{0.70\textsuperscript{\relsize{-3}***}} & 0.37\textsuperscript{\relsize{-3}***} & \cellcolor{green!25}\textbf{0.69\textsuperscript{\relsize{-3}***}} & 0.46\textsuperscript{\relsize{-3}***} \\
Dismissing Emotions & 0.35\textsuperscript{\relsize{-3}*} & 0.51\textsuperscript{\relsize{-3}**} & 0.49\textsuperscript{\relsize{-3}**} & 0.12 & 0.17\textsuperscript{\relsize{-3}**} & 0.10 \\ \bottomrule
\end{tabular}}
\caption{\mybold{Inter-Rater Reliability across Annotator Pairs, Sub-Components, and Frameworks.}
Inter-rater reliability (quadratically weighted $\kappa_w$)) for empathic communication across annotator pairs for each sub-component grouped by framework. Experts comparisons with the LLM (Gemini 2.5 Pro) and crowd compare median expert annotations with the LLM and crowd, respectively.
Kappa values above the high-agreement threshold (median expert agreement where $\kappa_w \geq 0.58$) are highlighted in \colorbox{green!25}{green}.
Statistical significance of $\kappa_w$ was assessed by testing the null hypothesis $H_0: \kappa_w = 0$ using a two-tailed Z-test. P-values are indicated by \textsuperscript{\relsize{-3}*} ($p < 0.05$), \textsuperscript{\relsize{-3}**} ($p < 0.01$), and \textsuperscript{\relsize{-3}***} ($p < 0.001$).}
\label{tab:kappa_table} 
\end{table}

\subsection*{Classification metrics obscure performance on subjective annotation tasks}

\begin{figure}[h!]
    \centering
    \includegraphics[width=\linewidth]{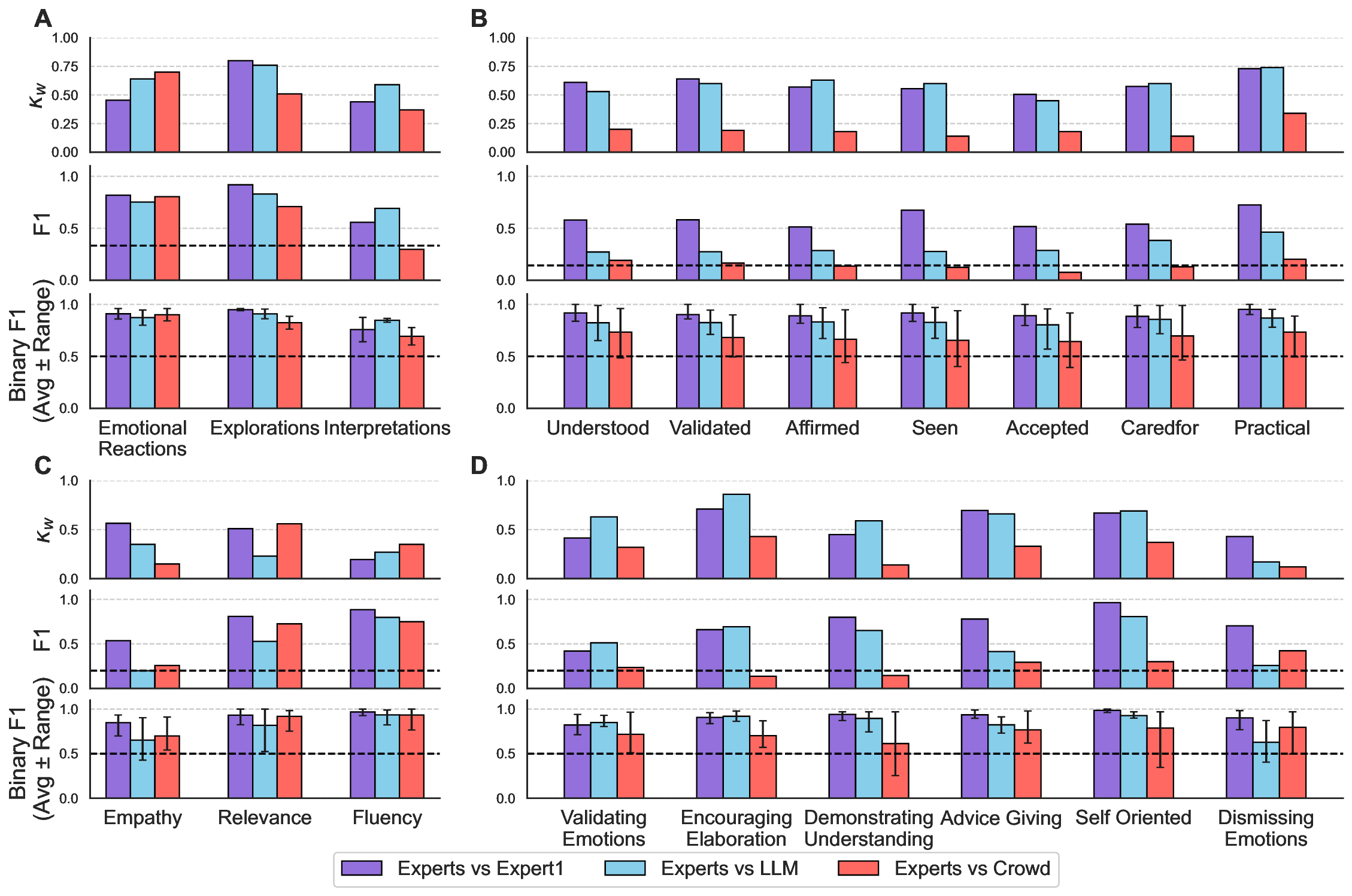}
    \caption{\textbf{Comparing Contextualized Inter-Rater Reliability with Multi-Class and Binary F1 Scores.} Experts vs Expert1 (\colorbox{ExpertColor!55}{purple}), Experts vs LLM (\colorbox{LLMColor!55}{(blue)}), and Experts vs Crowd (\colorbox{CrowdColor!55}{red}) present agreement metrics between the median expert annotation and expert 1, Gemini 2.5 Pro, and the crowd.  We present results for four datasets (A) EPITOME, (B) Perceived Empathy, (C) Empathetic Dialogues, and (D) Lend an Ear. Random classifier baselines are indicated by dotted (multi-class) and dashed (binary) lines.}
    \label{fig:f1-kappa-main}
\end{figure}

Following norms in the LLM-as-judge literature where median expert annotations are treated as ground truth and performance is evaluated as a classification task~\cite{gabriel2024can, yang2024social, ziems2024can, zhang2023wider}, we report F1 scores and compare them to contextualized Cohen's Kappa values. Figure \ref{fig:f1-kappa-main} presents Cohen's kappa values, multi-class F1 scores, and the average and range of binary F1 scores derived by selecting a single threshold to split the rating scale, and calculated across all possible thresholds for the datasets using median expert annotations as the ground truth. While F1 scores are useful for incorporating the trade-offs of false positives and false negatives when reporting classification performance, F1-scores for subjective, multi-class annotations face several limitations. 

First, classification metrics can obscure nuanced performance differences. On one hand, F1 scores can mask moderate inter-rater reliability when classes are highly imbalanced. A high F1 score can be achieved by accurately predicting a disproportionately large majority class, while agreement on minority classes remains poor like in the case of  ``Emotional Reactions'' from EPITOME. On the other hand, low F1 scores may conceal relatively high inter-rater reliability because classification metrics rely on a single ground truth and thus miss important nuances like off-by-one errors~\cite{groh2022towards}. For example, Figure~\ref{fig:f1-kappa-main} shows relatively high ($\kappa_w$) for expert agreement on ``Explorations'' in EPITOME and ``Practical Advice'' in Perceived Empathy, yet this relative performance compared to the other sub-components is not mirrored in corresponding multi-class F1 scores because F1 penalizes any deviation from exact label matches.

Second, F1 scores are sensitive to the rating scale. The F1 score random guessing baseline varies inversely with the number of categories, which complicates comparisons between annotation scales. For example, for experts, we find average F1 scores of 63.9 for sub-components in EPITOME (3 categories) but only 32.0 for Perceived Empathy (7 categories) reflecting sensitivity to rating scales rather than actual accuracy differences. 

Third, binary F1 scores depend on the threshold used to binarize a scale for evaluation. As demonstrated in Figure~\ref{fig:f1-kappa-main} (bottom panel, error bars represent the range of binary F1 scores), binary F1 scores calculated from the same underlying data can show drastic variation depending on the threshold for splitting the scale (e.g., 1 vs 2-7, or 1-3 vs 4-7, etc.). For instance, the Experts vs Crowd comparison for the ``Accepted'' sub-component in Perceived Empathy shows a low multi-class agreement (16.4 F1) which contrasts sharply with a high binary F1 score (68.4 F1, min:45, max: 94.7). Moreover, depending on how the threshold is selected for calculating a binary F1-score, all sub-components of the Perceived Empathy framework have binary F1 scores of 100\% and two of the three EPITOME sub-components have above 90\% binary F1 scores despite the majority of ($\kappa_w$) values between 0.4 and 0.6.

\subsection*{Crowd annotations distort evaluations of empathy}

Crowd annotations consistently show low reliability with expert annotations.
Across all frameworks, inter-rater reliability ($\kappa_w$) between crowd workers' annotations and expert annotations was between $0.1$ to $0.63$ (median = $0.33$), exceeding the high agreement threshold in only one sub-component. Appendix Figure~\ref{fig:crowd-empathy-inflation}  illustrates the distribution of annotations of crowds, experts, and LLMs for the four datasets.

Additionally, crowd workers' annotations are positively biased relative to experts' and LLMs' annotation. They assigned higher average ratings relative to experts on 18 out of the 21 sub-components evaluated. This systematic positive bias by crowdworkers reveals empathy inflation in novice perceptions of empathic support relative to trained experts. Lay annotators may rely on intuitive heuristics such as ``it's the thought that counts'', giving speakers the benefit of the doubt by focusing on perceived intentions rather than the content of the message. Such heuristic-driven evaluation, where annotators substitute complex judgments with simpler affect-based evaluations has been reported in crowdsourced emotional message labeling tasks~\cite{huffaker2020crowdsourced}.

Moreover, previous work finds that most people typically generate empathic messages with moderate, rather than high or low empathy when evaluated against theoretical frameworks~\cite{macgeorge2017influence}. If crowd annotators are unable to produce highly empathic messages, they may also fail to recognize or appreciate the value of those messages when others generate them.

\section*{Discussion}\label{discussion}
Our study demonstrates that LLMs can judge the nuances of empathic communication in text-based conversational contexts with a level of reliability approaching that of trained human experts (median expert-LLM $\kappa_w = 0.60$). While empathic communication is a context-dependent, subjective skill, that is difficult to objectively measure, many evaluative frameworks have been developed for teaching and measuring it in business~\cite{covey20207, drollinger2006development}, medicine~\cite{suchman1997model, mercer2004consultation, bylund2005examining}, communications~\cite{burleson2003emotional}, natural language processing of conversations~\cite{rashkin2018towards, sharma2020computational}, and psychology~\cite{yin2024ai}. When looking at the same conversations, experts generally agree but sometimes disagree with each other on how they would answer questions from these evaluative frameworks. By comparing 3,150 expert annotations, 2,844 crowd worker annotations, and 3,150 LLM annotations on four evaluative frameworks of empathic communication applied to 200 text-based conversations, we show that LLMs are generally reliable for judging empathic communication in the same conversational contexts and frameworks as when experts are reliable. 

Our evaluation draws from a wide range of conversational contexts across the four datasets we analyzed. The 200 conversations spanned everyday challenges such as workplace setbacks, financial strain, family conflict, and socially awkward incidents, as well as highly sensitive disclosures involving mental health struggles, self-harm, and experiences of bias or discrimination (Appendix Table \ref{suptab:interpretations}). These diverse themes ensured that our annotations covered many real-world contexts in which empathic communication plays a critical role. Specifically, we use four evaluative frameworks of empathic communication each corresponding to one of the datasets, to evaluate the conversations. Although each framework defines its own set of sub-components, some sub-components display clear conceptual overlap. For example, EPITOME's ``explorations'' and Lend an Ear's ``encouraging elaboration'' both focus on asking questions to probe the support-seeker's experiences. Similarly, Perceived Empathy's ``affirmed'' and EPITOME's ``emotional reactions'' target affirming the support-seeker.

Expert reliability varied depending on how constructs were operationalized within each framework. For example, constructs related to demonstrating understanding of the seeker’s situation (e.g., ``empathy'' in  Empathetic Dialogues, ``interpretations'' in EPITOME, and ``demonstrating understanding'' in Lend an Ear) exhibit low reliability. These sub-components lack explicit, observable cues, which introduces ambiguity. In contrast, constructs such as the ones focused on asking questions (e.g., in EPITOME and Lend an Ear) achieve high reliability, demonstrating conceptual clarity across conversational contexts. This variability echoes longstanding practices in qualitative coding and survey design~\cite{boyatzis1998transforming}, where codebooks are iteratively refined to emphasize observable cues rather than relying on subjective judgments alone. The same lesson applies to the growing use of LLMs-as-judge. If annotation questions are vague, redundant, or require subjective inferences, we should expect unreliable evaluations from both humans and models.  

The reliability of expert annotations reveals important strengths and limitations of each of the frameworks under investigation. EPITOME provides a relatively precise operationalization of ``explorations'' which showed the highest expert inter-rater reliability, but its other sub-components are ambiguous and difficult to annotate consistently. The Perceived Empathy framework showed relatively high reliability across sub-components (e.g., ``understood'', ``accepted'', ``cared for'', ``seen''), but many of these have high multicollinearity, which means they are redundant. In contrast, the Empathetic Dialogues framework is overly broad, collapsing all empathic behaviors into a single ``empathy'' sub-component, which limits its usefulness. The Lend an Ear framework was designed to address these issues by specifying six conceptually distinct sub-components. While it improves reliability overall,  experts still struggled with the ``dismissing emotions'' and ``demonstrating understanding'' sub-components suggesting that these require further refinement to be annotated reliably.

In addition to guiding data annotation, these frameworks have been used to train and evaluate models for a range of computational empathy tasks~\cite{sharma2020computational, gabriel2024can, ziems2024can} and to inform the design of AI conversational companions~\cite{sharma2023human, roller2020recipes}. Training or evaluating models on poorly specified, ambiguous, or redundant constructs may yield unreliable outputs. It is therefore important to critically assess each sub-component before incorporating it into pipelines for downstream applications. 

Our findings reveal that crowd workers' annotations are less aligned with experts' annotations than LLM annotations are with experts' annotations. We speculate that disparities between the crowd workers' and experts' annotations may arise for a number of reasons. First, it is possible that crowd workers and experts expended different levels of effort. Second, crowdworkers may have relied on intuitive heuristics and perceived intentions more than experts who experienced at focusing on a strict interpretation of the evaluative frameworks. Neither training crowdworkers nor aggregating their responses together addresses the gap in reliability between crowdworkers and experts. 

While LLMs' annotations match experts' more closely than crowdworkers' annotations, we find some superficial evidence that crowdworkers' ratings of empathic quality are more closely aligned with support-seekers’ self-reported perceptions of empathy. Specifically, for six of the nine sub-components in the Perceived Empathy dataset where self-report data are available, crowd annotations align more closely with participants' ratings of perceived empathy (median $\kappa_w$ = 0.33) compared to  expert annotations (median $\kappa_w$ = 0.11). However, this apparent alignment is an arifact of systematic rating inflation as both support-seekers' and crowdworkers' responses tend to cluster at the higher end of the Likert scale. This pattern is consistent with social desirability bias, creating artificial similarity in ratings rather than meaningful overlap in their assessments of empathic ability. Careful analysis using Kendall's $\tau_B$, which better accounts for ties and focuses on whether two groups order responses similarly, illustrates that the alignment is driven by these inflated ratings rather than an ordinal difference (see Appendix Figure~\ref{fig:selfvsothers}). Furthermore, support-seekers in this dataset often preferred AI-generated responses over human-written ones~\cite{yin2024ai}, suggesting that the qualities they prefer in a supportive response may be more nuanced than what is captured by their immediate self-reflection or by crowd evaluations. 

Inter-rater reliability relative to experts offers contextualized insights unavailable in evaluations where traditional classification metrics are applied to a label treated as the objective ground truth. The nature of evaluating subjective, context-dependent tasks involves addressing three inter-related questions: First, how comprehensive and reliable is a framework for measuring performance in a particular context? Second, how reliable are human experts with each other on a particular framework? Third, how reliable are LLMs relative to experts? If human experts have low inter-rater reliability on a task, then the task may simply be too ambiguous and need further refinement. On the other hand, if human experts demonstrate high inter-rater reliability, then there's evidence that justified agreement is possible. By reporting LLMs' inter-rater reliability with experts alongside expert agreement on multiple constructs, it is possible to offer a comprehensive report on construct reliability, expert performance, and generalizable LLM alignment with expert performance. In contrast, traditional classification metrics leave us with blind spots. First, classification metrics obscure performance. High F1-scores mask relatively moderate inter-rater reliability when classes are unbalanced. Likewise, low F1-scores can mask relatively high inter-rater reliability when there are many off by one instances. Second, classification metrics are sensitive to the number of categories in the rating scale, which makes cross framework evaluations difficult. Third, binarization of classification metrics can lead to a large range of possible scores to present, which can be either misleading if a single arbitrary cutoff for binarization is chosen or hard to interpret if the full range is shown. These blindspots echo research in affective computing which argues that subjective, context‑dependent constructs like emotion are better captured through ordinal or graded judgments that emphasize relative comparisons, rather than forcing them into categorical labels~\cite{yannakakis2017ordinal}.

We offer several recommendations for practitioners and researchers aiming to evaluate subjective constructs reliably at scale. First, our results highlight the importance of benchmarking expert performance on evaluative frameworks of subjective constructs to contextualize the levels of subjectivity of sub-components within the frameworks. Second, this empirical benchmark of expert agreement can guide subsequent refinement of evaluative frameworks. Third, LLMs should be measured against the reliability of experts to offer a comprehensive evaluation and avoid the blind spots that traditional classification metrics and assumptions of an objective ground truth tend to obscure. Finally, domain experts can be helpful for extending evaluative frameworks into well-structured prompts for LLMs to maximize LLMs' alignment with experts on a LLM-as-judge task. Adopting these practices can enhance the reliability of large-scale subjective annotations by LLMs.

\subsection*{Limitations}\label{Limitations}

We specifically focus on interactions between strangers, which serves as a useful starting context for many LLM-as-companion applications. Future work should look at when and how evaluations of empathic communication may generalize (or not) to more complex socioaffective scenarios, such as longitudinal interactions or emotionally intensive relationships~\cite{kirk2025human}.

In order to evaluate the reliability of LLMs at judging empathic communication, we developed an evaluation framework that allows for direct comparisons across experts, crowds, and LLMs. In particular, each annotator group was asked the same straightforward questions outlined in Figure~\ref{fig:stacked_overview} for each annotation with minimal or no training. This approach is essential for keeping the task tractable for crowdworkers~\cite{aroyo2015truth}. However, our approach does not match the standard practice in communications research where experts build an explicit coding framework and coders typically undergo iterative training until they reach a minimum threshold of inter-rater reliability ~\cite{skjott2019coding, samter2016coding}. Therefore, the expert reliability reported here represents a conservative lower bound on the achievable expert reliability in the absence of further structured coding. Likewise, further prompt refinements may further increase alignment between LLMs and experts.

\subsection*{Conclusion}\label{Conclusion}

We evaluated the reliability of large language models relative to experts and crowd annotators at judging nuances of empathic communication across multiple evaluative frameworks and conversational datasets. Our analysis revealed that LLMs are generally reliable for judging empathic communication in the same contexts as when experts are reliable. Moreover, LLMs' annotations are much more reliable than those of crowdworkers. LLM-expert reliability was consistently higher than crowd-expert reliability across all sub-components where experts had substantial agreement. Expert agreement was strongest for sub-components with clear, observable behaviors and weaker for more subjective constructs. The context of expert agreement offered insights otherwise obscured by evaluations based on classification metrics where annotations are operationalized on the assumption of an objective ground truth. Overall, this work offers a foundation for more robust benchmarking of subjective NLP tasks and points toward the possibility of LLMs-as-judge for transparency and accountability when deploying LLMs-as-companions for socially-sensitive applications.

\section*{Methods}\label{methods}

We compare the inter-rater reliability of annotations by experts, crowds, and LLMs across evaluative frameworks of empathic communication applied to four conversational datasets. Specifically, we analyze 3,150 annotations from experts, 2,844 annotations from crowd workers, and 3,150 annotations from LLMs across the 21 sub-components of the four evaluative frameworks. 

\subsection*{Evaluative Frameworks and Accompanying Conversational Datasets}

We examine annotations of empathic communication across four evaluative frameworks and accompanying conversational datasets: Empathetic Dialogues~\cite{rashkin2018towards}, EPITOME~\cite{sharma2020computational}, Perceived Empathy\cite{yin2024ai}, and the Lend an Ear pilot. Each dataset contains English-language conversations where one participant shares a personal challenge, and the other provides support. Table~\ref{tab:sub_context_comparison} summarizes the contextual and structural differences between the datasets and Figure~\ref{fig:stacked_overview} lists the exact annotation questions of each framework. 

In order to make expert annotations tractable, we sampled a total of 200 conversations, with 50 from each dataset. Crowd workers previously annotated the three published datasets~\cite{rashkin2018towards, sharma2020computational, yin2024ai}, assigning empathy scores across multiple sub-components. We stratified our random samples from the four datasets based on these crowdsourced annotations, selecting 10 from the highest-scoring quartile, 10 from the lowest, and 30 from the middle range. For the Lend an Ear dataset, we randomly sampled 50 of the 150 conversations from the pilot.

\subsubsection*{Empathetic Dialogues}
The Empathetic Dialogues conversational dataset contains 24,850 crowdsourced dyadic conversations collected via the ParlAI platform~\cite{miller2017parlai, rashkin2018towards}, with 810 Amazon Mechanical Turk workers alternating between speaker and listener roles. Speakers described situations based on one of 32 emotion labels (e.g., proud, sad, anxious), followed by a conversation with a listener who was unaware of the assigned label. Focusing on conversations that call for empathetic support, we filtered for conversations prompted with negative emotions including sadness, annoyance, fear, anxiety, guilt, disappointment, embarrassment, and shame and then we randomly sampled from these conversations. Two crowd annotators rated each conversation on a 5-point Likert scale (1 = not at all, 5 = very much) across three sub-components. The original study~\cite{rashkin2018towards} does not specify any training details for annotators.

\subsubsection*{EPITOME}
The EPITOME conversational dataset contains 10,143 post-response pairs from two data sources: 7,062 pairs from TalkLife and 3,081 pairs from mental health subreddits on Reddit~\cite{sharma2020computational}. For this study, we sampled data from the open-sourced Reddit subset. Each pair was labeled using the EPITOME (EmPathy In Text-based asynchrOnous MEntal health) framework, which categorizes empathy for the domain of text-based, synchronous mental health conversations into three sub-components: Emotional Reactions, Interpretations, and Explorations. Eight Upwork annotators, trained for up to an hour with feedback on 100 annotations and spot checks, evaluated responses using the EPITOME framework on a 3-point scale: no, weak, or strong communication of each sub-component of empathy. Each response is evaluate by a single annotator.

\subsubsection*{Perceived Empathy}
The Perceived Empathy conversational dataset contains two-turn conversations collected in stages~\cite{yin2024ai}. In the first stage, 501 participants recruited from Prolific shared a complex personal situation via voice recording. Next, 233 additional Prolific workers respond in writing to one of the 501 shared situations and Bing Chat AI generated responses for the rest of the shared situations. We filtered for conversations where responses were provided by humans. Finally, participants were asked to rate how heard the responses made them feel. In a follow-up study, 1,449 annotators evaluated the two turn conversations from the main study based on the degree to which the response would make the discloser feel understood, affirmed, validated, seen, accepted, and cared for, the degree to which the response provided emotional or practical support, and the effort the responder put into into crafting their reply. 
Annotators provided ratings on a 7-point Likert scale (1 = not at all, 7 = very much) and each response in the dataset was rated by 2 to 4 annotators. The mean of their ratings was used as the final annotation. 

\subsubsection*{Lend an Ear Pilot}
The Lend an Ear pilot dataset contains 150 conversations from 50 participants recruited from Prolific. Each participant engaged in three four-minute conversations where they provided empathetic support to a partner sharing a workplace concern. From this dataset, we randomly sampled 50 conversations. We recruited 150 Prolific annotators to evaluate responses across multiple empathy sub-components, including the extent to which the participant validated the speaker's emotions, encouraged further elaboration through open-ended questions, demonstrated understanding of the speaker's issues, offered unsolicited advice, shifted the focus to themselves, or dismissed/minimized the speaker's feelings. Annotators were provided with examples for each sub-component to calibrate their evaluations. Each annotator rated responses on a 5-point Likert scale (1 = not at all, 5 = very much). Each conversation in the dataset was evaluated by between 2 and 11 annotators. The final annotation for each conversation was computed by averaging the annotators' ratings.  We include the exact scenarios and instructions provided to participants in the Appendix.

\subsection*{Conversational Contexts Across Datasets}
We trained sparse autoencoders (SAEs) on 200 conversations across the EPITOME, Empathetic Dialogues, Perceived Empathy, and Lend an Ear pilot datasets to categorize the themes of support seekers' personal disclosures~\cite{singh2025discovering, peng2025use}. Based on power-of-two grid search of the number of topics from $2^{3}=8$ to $2^{6}=64$, we evaluated silhouette scores and qualitatively examined the resulting themes with their corresponding conversations. This analysis indicated that 16 topics provided the most informative representation of conversational themes. 

The conversations reflect a fairly broad set of everyday concerns that arise in empathic exchanges. Appendix Table~\ref{suptab:interpretations} summarizes the distribution of SAE-identified topics which range from mental health challenges (e.g., depression, suicidal ideation, self-harm), to family- and work-related struggles (e.g., losing a job, being passed over for a promotion, conflict with relatives), to personal stressors such as financial strain or socially awkward incidents. 

Topic coverage is not evenly distributed across datasets. In the EPITOME dataset, conversations focus on mental health struggles and explicit expressions of self-harm. In contrast, the Perceived Empathy and Empathetic Dialogues datasets encompass a broader range of personal, relational, financial, and social concerns. The Lend an Ear pilot dataset revolves around three workplace challenges: feeling overworked, losing a job, and getting passed over for promotion. The passed over for promotion scenario includes a detailed backstory addressing issues that may come up in conversations by people from historically marginalized backgrounds (see Appendix for detailed scenario prompts). In 3 out of 19 conversations, the LLM-based support seeker explicitly mentioned suspecting racial bias as a factor in their lack of advancement.

\subsection*{Expert Annotations}
Three of the authors on this paper independently provided 1050 annotations each. One annotator is a senior communications professor who has led hundreds of workshops on empathic communication. Another is a junior scholar who has trained with this first senior communications professor and spent over a year reviewing frameworks for empathic communication. The third annotator is another senior communications professor who has published many peer reviewed papers on social support, interpersonal skills, and advice giving. For clarity and conciseness, we refer to these individuals simply as experts. Our analysis reveals these experts mostly agree on annotations but sometimes disagree. These reasonable disagreements offer evidence that expert annotations should not be treated as ground truth upon which to evaluate model performance with classification metrics. Instead, the appropriate comparison is how the inter-rater reliability between the three experts' annotations compares to the inter-rater reliability between median experts' annotations and the LLMs' and crowds' annotations following~\cite{groh2022towards}.

In the annotation process, experts were provided with a brief overview of the frameworks and datasets and instructed to evaluate the conversations using the same criteria, rating scale, and questions provided to the crowdsourced annotators, as outlined in the respective dataset descriptions. In total, each expert provided 1050 annotations across four datasets. We use the median expert ratings (calculated from the three individual expert annotations) as the reference expert annotation for comparison with LLM and crowd-sourced annotations. 

\subsection*{Crowd Annotations}
We used the mean of crowdsourced annotations as the final annotation by the crowd for each framework. For the Empathetic Dialogues, EPITOME, and Perceived Empathy datasets, we relied on the crowd annotations provided by the original authors. In the Empathetic Dialogues dataset, two independent annotators annotated each conversation, and their ratings were averaged to produce a single score. In the EPITOME dataset, there was only one annotation per conversation. For the Perceived Empathy dataset, we used the mean ratings provided by the authors, which were provided by 2-4 annotators per conversation. For the Lend an Ear pilot experiment, we crowdsourced annotations, with each conversation being evaluated by between 2 and 11 independent annotators per conversation. We used mean crowd ratings for all analyses to ensure consistency across datasets since only means were available for Perceived Empathy and EPITOME had a single annotator.

\subsection*{LLM Annotations}
In order to generate annotations, we systematically prompt the LLM based on three criteria. First, the prompt included a framework for empathic communication grounded in our interpretation of the communications and psychology literature on empathic communication. See Supplemental Information for specific wording. 

Second, we included three-shot examples with expert provided scores specific to each dataset to illustrate the evaluation criteria and expected output format. This second criteria follows the HELM methodology for including in-context examples that increases the likelihood that LLMs generate a response that fits within evaluation scale~\cite{liang2022holistic}. Third, we input the conversation to be assessed and instructed the LLM to assign scores using the Likert scale specific to each task. 

The results presented in this paper are based on annotations generated by Gemini (gemini-2.5-pro-preview-03-25). Appendix Table \ref{ED:llms-kappa-median} shows reliability of annotations generated by GPT-4 (gpt-4o-2024-08-06) and Claude (claude-3-7-sonnet-20250219). All LLM API calls were made with a temperature setting of zero to ensure consistent outputs.

\subsection*{Measuring Inter-Rater Reliability}
We assess annotation quality across different frameworks using two complementary statistical measures: inter-rater reliability via Cohen’s Kappa, and classification accuracy using F1 scores, common in supervised machine learning contexts. These metrics highlight distinct aspects of annotator agreement and variability. We primarily focus on weighted Cohen’s Kappa ($\kappa_w$) which takes into account the degree of disagreement between annotators, penalizing differences by a squared term~\cite{warrens2015five}. Specifically, we compute $\kappa_w$ for each expert pair, and between the median expert annotation and annotations from (a) the LLM and (b) crowd workers. We also  calculate Krippendorff’s $\alpha$, which generalizes to any number of annotators, for experts and LLM annotations.

While our primary analysis relies on inter-rater reliability metrics, we also report multi-class and binary F1 scores treating median expert annotations as ground truth. We do this to contrast our approach with previous work on LLM-as-judge which treats median expert annotations as ground truth and relies on classification metrics to report annotation performance~\cite{openai2023contentmoderation, gu2024survey, ziems2024can}. We demonstrate how classification metrics fall short in providing meaningful insights for subjective tasks compared to reliability metrics.

\subsection*{Qualitative Evaluation}
In an effort to understand the differences in annotations across experts, crowds, and LLMs, we conducted a qualitative analysis of annotation patterns across datasets and annotator groups.
After gathering all expert annotations, we systematically filtered for those conversations in which experts disagreed with one another as well as those in which expert and crowd judgments diverged. Because only the Lend an Ear and EPITOME datasets included annotation explanations from both experts and crowd workers, our qualitative analysis focuses on these two conversational datasets. For each conversation where annotators differed, one expert reviewed all expert and crowd explanations and highlighted reasons to help explain why annotator may have arrived at different ratings. Finally, all experts discussed five representative conversations in depth and reviewed the qualitative insights presented in the Appendix. 

\section*{Ethics Approval and Consent to Participate}
This research complied with all relevant ethical regulations and obtained informed consent from all participants for data we collected. The Northwestern University Institutional Review Board (IRB) determined that the research met the criteria for exemption from further review. The study's IRB identification number is STU00222032 and STU00223043.

\subsection*{Ethical Implications of Empathy Evaluation}\label{Ethics and Safety}

The integration of LLMs in evaluating and expressing empathic support raises important ethical considerations. Empathic communication is a critical interpersonal skill that affects both people's lives and their livelihoods.

LLMs offer a promising path toward two important use cases including supporting skill training by scaling evaluations and providing accessible feedback, and augmenting AI companions by surfacing harmful communication patterns in human-AI conversations and enabling more responsive and sensitive interactions. But there remain important risks. Unreliable or biased evaluations could propagate poor practices and harm the people these systems are meant to serve. For example, unchecked empathic responses by AI companions can foster unhealthy attachments or emotionally manipulative tactics~\cite{de2025unregulated, de2025emotional}.

Our study offers a step towards understanding the differences between experts, crowds, and LLM evaluation of empathic communication. While LLMs have higher inter-rater reliability with expert annotations relative to the crowd, the variability in expert reliability across frameworks reveals the complexity of evaluation and the need to carefully test and validate frameworks before using LLMs in annotation pipelines. Future work should also consider domain‑specific guidelines for acceptable error rates before deploying automated assessments into professional training and evaluation of AI companions.

\section*{Data availability}
The data used include datasets from Empathetic Dialogues~\cite{rashkin2018towards}, EPITOME~\cite{sharma2020computational}, Perceived Empathy~\cite{yin2024ai}, and the Lend an Ear pilot. The full data (except the Perceived Empathy conversational dataset) used during the current study are available in the data folder in our public GitHub repository: \href{https://github.com/aakriti1kumar/replication-data-and-code-when-LLMs-reliable-empathic-communication}{https://github.com/aakriti1kumar/replication-data-and-code-when-LLMs-reliable-empathic-communication}. The full Perceived Empathy dataset is available upon request to the authors of the associated paper.

\section*{Code availability}
The code used during the current study are available in a public GitHub repository: \href{https://github.com/aakriti1kumar/replication-data-and-code-when-LLMs-reliable-empathic-communication}{https://github.com/aakriti1kumar/replication-data-and-code-when-LLMs-reliable-empathic-communication} and archived at Zenodo~\cite{Kumar2025LLMReplication}.

\section*{Author Contributions}

A.K. and M.G conceived the investigation, A.K. curated the data, N.P., E.F., B.L. annotated the data, A.K., N.P., and M.G. analyzed the results, A.K. and M.G. wrote the initial manuscript, A.K., N.P., D.Y., E.F., B.L., M.G. reviewed and edited the manuscript.

\section*{Competing Interests}
The authors declare no competing interests.

\begin{appendices}

\bibliography{sn-bibliography}
\newpage
\section{LLM-Expert Inter-Rater Reliability under Different Prompting Approaches}\label{ED:llm-sec}

\newcolumntype{L}{>{\hsize=1.2\hsize\raggedright\arraybackslash}X}
\newcolumntype{C}{>{\centering\arraybackslash}X}

\begin{table}[h!]
\centering
\begin{tabularx}{\textwidth}{|L|*{6}{C|}}
\toprule
\textbf{Sub-component} & \textbf{Zero-shot} & \textbf{Few-shot} & \textbf{Framework} & \textbf{Framework + Few-shot} \\
\midrule
\multicolumn{5}{l}{\textbf{Empathetic Dialogues}} \\
\cmidrule{1-5}
Empathy & 0.44\textsuperscript{\relsize{-3}**} & 0.36\textsuperscript{\relsize{-3}***} & 0.46\textsuperscript{\relsize{-3}***} & 0.35\textsuperscript{\relsize{-3}***} \\
Fluency & \cellcolor{red!25}-0.05 & 0.08 & 0.23 & 0.27 \\
Relevance & \cellcolor{green!25}\textbf{0.60\textsuperscript{\relsize{-3}**}} & 0.27 & \cellcolor{green!25}\textbf{0.61\textsuperscript{\relsize{-3}**}} & 0.23 \\
\midrule
\multicolumn{5}{l}{\textbf{EPITOME}} \\
\cmidrule{1-5}
Emotional Reactions & 0.25\textsuperscript{\relsize{-3}***} & 0.53\textsuperscript{\relsize{-3}***} & 0.22\textsuperscript{\relsize{-3}***} & \cellcolor{green!25}\textbf{0.64\textsuperscript{\relsize{-3}***}} \\
Explorations & 0.32\textsuperscript{\relsize{-3}*} & 0.54\textsuperscript{\relsize{-3}***} & 0.30\textsuperscript{\relsize{-3}*} & \cellcolor{green!25}\textbf{0.76\textsuperscript{\relsize{-3}***}} \\
Interpretations & 0.53\textsuperscript{\relsize{-3}***} & 0.56\textsuperscript{\relsize{-3}***} & 0.51\textsuperscript{\relsize{-3}***} & \cellcolor{green!25}\textbf{0.59\textsuperscript{\relsize{-3}***}} \\
\midrule
\multicolumn{5}{l}{\textbf{Perceived Empathy}} \\
\cmidrule{1-5}
Understood & 0.55\textsuperscript{\relsize{-3}***} & \cellcolor{green!25}\textbf{0.70\textsuperscript{\relsize{-3}***}} & 0.41\textsuperscript{\relsize{-3}***} & 0.53\textsuperscript{\relsize{-3}***} \\
Validated & 0.50\textsuperscript{\relsize{-3}***} & \cellcolor{green!25}\textbf{0.74\textsuperscript{\relsize{-3}***}} & 0.47\textsuperscript{\relsize{-3}***} & \cellcolor{green!25}\textbf{0.60\textsuperscript{\relsize{-3}***}} \\
Affirmed & 0.36\textsuperscript{\relsize{-3}**} & \cellcolor{green!25}\textbf{0.65\textsuperscript{\relsize{-3}***}} & 0.45\textsuperscript{\relsize{-3}***} & \cellcolor{green!25}\textbf{0.63\textsuperscript{\relsize{-3}***}} \\
Accepted & 0.29\textsuperscript{\relsize{-3}*} & 0.34\textsuperscript{\relsize{-3}*} & 0.34\textsuperscript{\relsize{-3}*} & 0.45\textsuperscript{\relsize{-3}***} \\
Cared For & 0.34\textsuperscript{\relsize{-3}**} & 0.55\textsuperscript{\relsize{-3}***} & 0.39\textsuperscript{\relsize{-3}**} & \cellcolor{green!25}\textbf{0.60\textsuperscript{\relsize{-3}***}} \\
Seen & 0.50\textsuperscript{\relsize{-3}***} & \cellcolor{green!25}\textbf{0.63\textsuperscript{\relsize{-3}***}} & 0.46\textsuperscript{\relsize{-3}***} & \cellcolor{green!25}\textbf{0.60\textsuperscript{\relsize{-3}***}} \\
Emotional & 0.46\textsuperscript{\relsize{-3}***} & \cellcolor{green!25}\textbf{0.69\textsuperscript{\relsize{-3}***}} & 0.49\textsuperscript{\relsize{-3}***} & \cellcolor{green!25}\textbf{0.75\textsuperscript{\relsize{-3}***}} \\
Practical & \cellcolor{green!25}\textbf{0.64\textsuperscript{\relsize{-3}***}} & \cellcolor{green!25}\textbf{0.71\textsuperscript{\relsize{-3}***}} & \cellcolor{green!25}\textbf{0.63\textsuperscript{\relsize{-3}***}} & \cellcolor{green!25}\textbf{0.74\textsuperscript{\relsize{-3}***}} \\
Motivation & 0.22\textsuperscript{\relsize{-3}**} & \cellcolor{green!25}\textbf{0.73\textsuperscript{\relsize{-3}***}} & 0.13 & \cellcolor{green!25}\textbf{0.70\textsuperscript{\relsize{-3}***}} \\
\midrule
\multicolumn{5}{l}{\textbf{Lend an Ear}} \\
\cmidrule{1-5}
Validating Emotions & 0.56\textsuperscript{\relsize{-3}***} & \cellcolor{green!25}\textbf{0.63\textsuperscript{\relsize{-3}***}} & \cellcolor{green!25}\textbf{0.59\textsuperscript{\relsize{-3}***}} & \cellcolor{green!25}\textbf{0.63\textsuperscript{\relsize{-3}***}} \\
Demonstrating Understanding & \cellcolor{green!25}\textbf{0.65\textsuperscript{\relsize{-3}***}} & 0.56\textsuperscript{\relsize{-3}***} & 0.46\textsuperscript{\relsize{-3}***} & \cellcolor{green!25}\textbf{0.59\textsuperscript{\relsize{-3}***}} \\
Encouraging Elaboration & \cellcolor{green!25}\textbf{0.80\textsuperscript{\relsize{-3}***}} & \cellcolor{green!25}\textbf{0.75\textsuperscript{\relsize{-3}***}} & \cellcolor{green!25}\textbf{0.80\textsuperscript{\relsize{-3}***}} & \cellcolor{green!25}\textbf{0.86\textsuperscript{\relsize{-3}***}} \\
Advice Giving & 0.51\textsuperscript{\relsize{-3}***} & 0.54\textsuperscript{\relsize{-3}***} & \cellcolor{green!25}\textbf{0.60\textsuperscript{\relsize{-3}***}} & \cellcolor{green!25}\textbf{0.66\textsuperscript{\relsize{-3}***}} \\
Self-Oriented & \cellcolor{green!25}\textbf{0.64\textsuperscript{\relsize{-3}***}} & \cellcolor{green!25}\textbf{0.64\textsuperscript{\relsize{-3}***}} & \cellcolor{green!25}\textbf{0.79\textsuperscript{\relsize{-3}***}} & \cellcolor{green!25}\textbf{0.69\textsuperscript{\relsize{-3}***}} \\
Dismissing Emotions & 0.31\textsuperscript{\relsize{-3}**} & 0.20\textsuperscript{\relsize{-3}**} & 0.35\textsuperscript{\relsize{-3}***} & 0.17\textsuperscript{\relsize{-3}**} \\ \bottomrule
\end{tabularx}
\captionsetup{labelformat=empty}
\caption{\textbf{Inter-Rater Reliability (Weighted Cohen's Kappa) of LLM Prompt Variants vs Expert Median across datasets and sub-components.} The prompt variants include \textbf{Zero-Shot}, \textbf{Few-Shot}, \textbf{Framework}, and \textbf{Framework + Few-Shot}, ordered by increasing complexity of the prompt for the LLM annotator (Gemini-2.5-pro-exp). Cells in \colorbox{green!25}{green} exceed the high-agreement threshold ($\kappa_w \geq 0.58$). Significance: * ($p < 0.05$), ** ($p < 0.01$), *** ($p < 0.001$).}
\label{ED:prompt-variants}
\end{table}

\clearpage
\section{Conversational Contexts Across Datasets}\label{ED:topics-conversation}

\begin{table}[ht]
\centering
\resizebox{\linewidth}{!}{%
\begin{tabular}{l c c c c c}
\toprule
\textbf{SAE-Identified Topic} & \textbf{All} &\makecell{\textbf{Empathetic} \\ \textbf{Dialogues}} & 
\textbf{EPITOME} & \makecell{\textbf{Perceived} \\ \textbf{Empathy}} & 
\makecell{\textbf{Lend an} \\ \textbf{Ear}} \\
\midrule
Mentions specific, non-serious incidents or events 
causing embarrassment or awkwardness in everyday situations & 19.5\% & 78\% & -- & -- & -- \\
Expresses explicit thoughts, plans, or desires related to suicide or self-harm & 18.0\% & -- & 70\% & 2\% & -- \\
Mentions being passed over for a promotion & 9.5\% & -- & -- & -- & 38\% \\
Mentions challenges in balancing work responsibilities following a promotion & 9.5\% & -- & -- & -- & 38\% \\
Mentions challenges or decisions related to housing, home ownership, or home repairs & 8.5\% & -- & -- & 34\% & -- \\
Mentions familial relationships or conflicts involving family members & 7.5\% & 2\% & -- & 28\% & -- \\
Mentions losing a job or being laid off & 6.0\% & -- & -- & -- & 24\% \\
Mentions work-related conflicts, challenges, or ethical dilemmas & 4.5\% & 2\% & -- & 16\% & -- \\
Mentions financial challenges or disputes involving family members or spouses & 4.5\% & -- & -- & 18\% & -- \\
Mentions experiences, symptoms, or struggles related to depression & 3.5\% & -- & 14\% & -- & -- \\
Mentions specific incidents or events causing annoyance or frustration in a personal context & 3.0\% & 12\% & -- & -- & -- \\
Mentions experiences, symptoms, or treatments related to mental health struggles or addiction & 2.0\% & -- & 6\% & 2\% & -- \\
Mentions jealousy or envy related to others' financial or material possessions & 1.5\% & 4\% & 2\% & -- & -- \\
Discusses personal experiences or questions related to antidepressant or antipsychotic medications and their effects & 1.5\% & -- & 6\% & -- & -- \\
Mentions personal aspirations or challenges related to physical or professional capabilities & 1.0\% & 2\% & 2\% & -- & -- \\
\bottomrule
\end{tabular}}
\captionsetup{labelformat=empty}
\caption{\textbf{Conversational contexts across datasets.} Sixteen themes of personal disclosures identified by sparse autoencoders (SAEs) across 200 conversations. Values represent the percentage of conversations within each dataset (column) in which the theme appeared.}
\label{suptab:interpretations}
\end{table}

\newpage
\section{Experts' Inter-Rater Reliability with Each Other}
\begin{figure}[htbp]
    \centering
    \includegraphics[width=.9\linewidth]{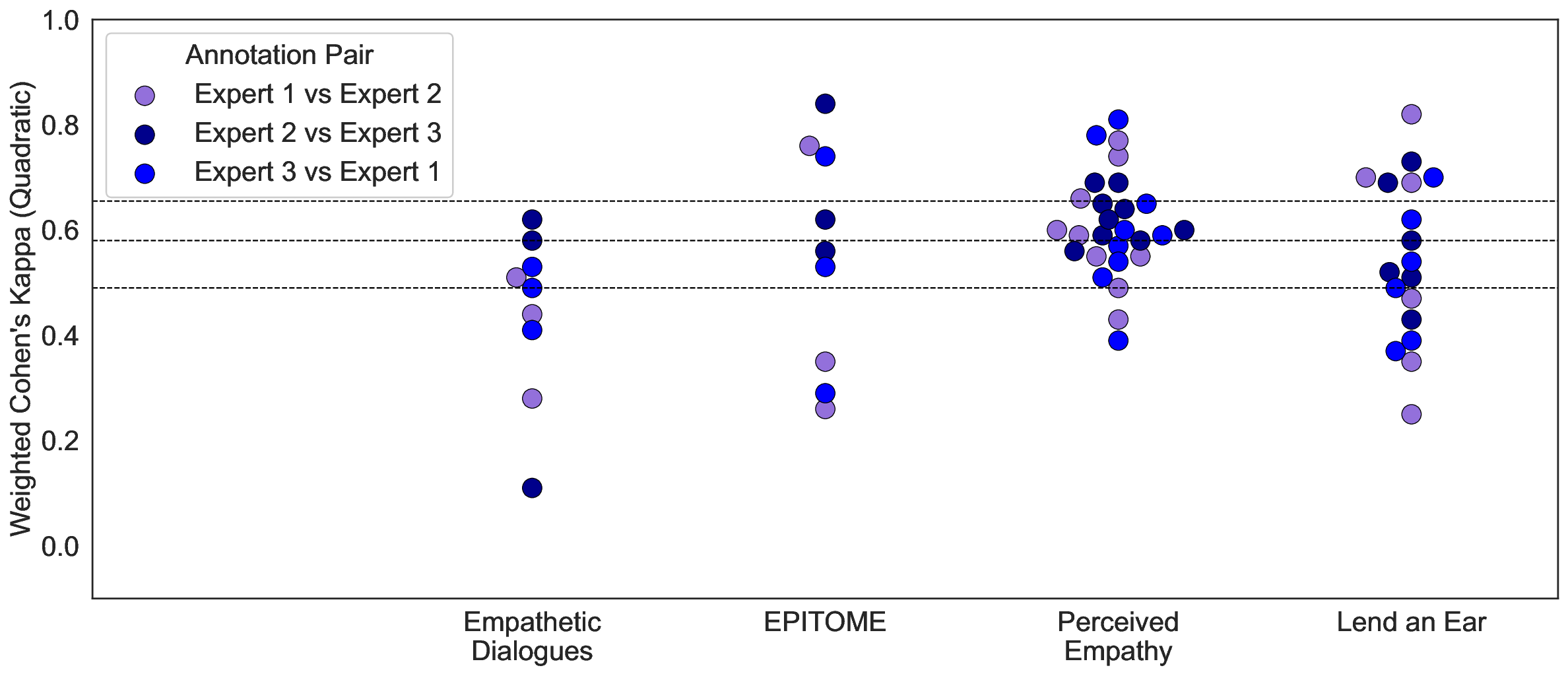}

    \caption{\textbf{Expert Inter-rater Reliability across Frameworks.} Pairwise expert reliability across four empathic communication frameworks}
    \label{fig:expertIRR}
\end{figure}

\begin{figure}[htbp]
    \centering
    \includegraphics[width=.9\linewidth]{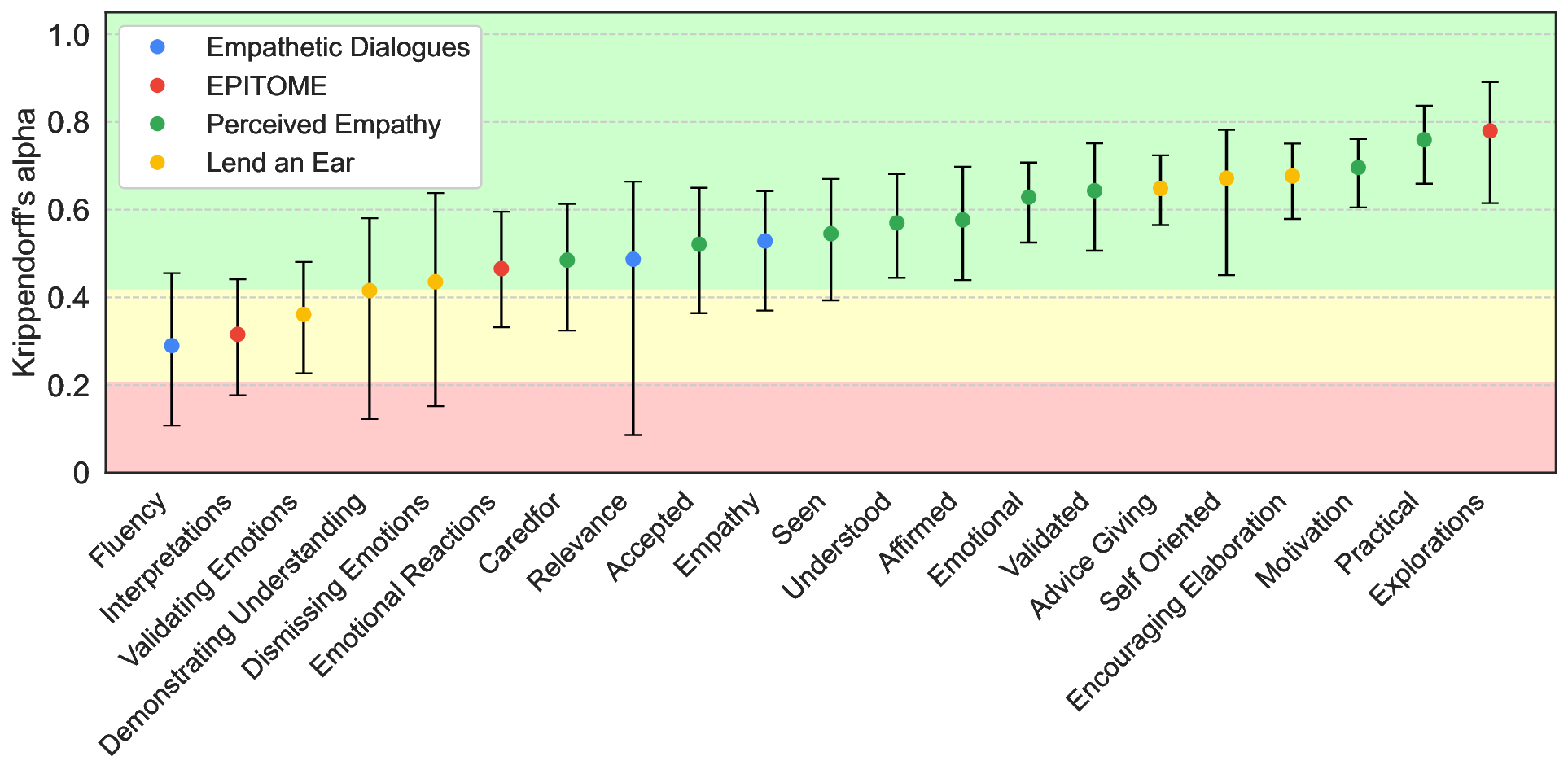}

    \caption{\textbf{Expert Inter-rater Reliability across Sub-components.} Inter-rater agreement as Krippendorff's alpha among the three expert annotators across frameworks' sub-components. Error bars are 95\% confidence intervals obtained by bootstrapping conversations, and the background color indicates the degree of agreement \cite{regier2013dsm}.}
    \label{fig:k-alpha}
\end{figure}

\newpage
\section{LLMs' Inter-Rater Reliability with Each Other}\label{ED:llm-sec}

\begin{table}[htbp]
    \centering
    \small
    \resizebox{\linewidth}{!}{%
    \begin{tabular}{@{}llrrrrrrrrr@{}}
    \toprule
        \multicolumn{2}{c}{} & 
        \multicolumn{1}{c}{\textbf{Experts}}& 
        \multicolumn{2}{c}{\textbf{Gemini (2.5 pro)}} & 
        \multicolumn{2}{c}{\textbf{OpenAI (GPT-4o)}} & 
        \multicolumn{2}{c}{\textbf{Claude (3.7 Sonnet)}} \\
        \cmidrule(lr){3-3}
        \cmidrule(lr){4-5} 
        \cmidrule(lr){6-7} 
        \cmidrule(lr){8-9}
    \textbf{Framework}   & \textbf{Sub-component}   & \textbf{Median} &\textbf{Fewshot} & \textbf{CoT}   & \textbf{Fewshot} & \textbf{CoT}    & \textbf{Fewshot} & \textbf{CoT}   \\
    \midrule
    \multirow{3}{*}{\makecell{Empathetic\\Dialogues}} & Empathy & 0.53 & 0.45 & 0.55 & 0.29 & 0.43 & 0.29 & 0.35\\
     & Relevance & 0.49 & 0.45 & 0.48 & 0.30 & 0.22 & 0.47 & 0.53 \\
     & Fluency & 0.28 & 0.44 & 0.48 & 0.15 & 0.20 & 0.21 & 0.45 \\
     \midrule
    \multirow{3}{*}{EPITOME} & Emotional Reactions & 0.53 & 0.46 & 0.47 & 0.67 & 0.50 & 0.71 & 0.58 \\
     & Explorations & 0.76 & 0.80 & 0.89 & 0.88 & 0.64 & 0.77 & 0.73 \\
     & Interpretations & 0.29 & 0.55 & 0.45 & 0.26 & 0.44 & 0.52 & 0.64 \\
     \midrule
    \multirow{9}{*}{\makecell{Perceived\\Empathy}} & Understood & 0.60 & 0.53 & 0.65 & 0.40 & 0.51 & 0.36 & 0.47 \\
     & Validated & 0.65 & 0.49 & 0.48 & 0.40 & 0.59 & 0.58 & 0.62 \\
     & Affirmed & 0.60 & 0.55 & 0.38 & 0.37 & 0.47 & 0.52 & 0.64\\
     & Seen & 0.55 & 0.55 & 0.43 & 0.47 & 0.49 & 0.37 & 0.61\\
     & Accepted & 0.57 & 0.37 & 0.38 & 0.23 & 0.49 & 0.47 & 0.57\\
     & Cared For & 0.55 & 0.47 & 0.35 & 0.34 & 0.29 & 0.53 & 0.32\\
     & Emotional & 0.59 & 0.71 & 0.69 & 0.70 & 0.69 & 0.65 & 0.75 \\
     & Practical & 0.77 & 0.69 & 0.69 & 0.62 & 0.67 & 0.62 & 0.73\\
     & Motivation & 0.66 & 0.68 & 0.77 & 0.47 & 0.67 & 0.73 & 0.80 \\
     \midrule
     
    \multirow{6}{*}{\makecell{Lend \\an Ear}} & Validate Emotions & 0.39 & 0.55 & 0.44 & 0.52 & 0.31 & 0.68 & 0.52 \\
     & Encouraging Elaboration & 0.69 & 0.80 & 0.64 & 0.70 & 0.49 & 0.79 & 0.67 \\
     & Demonstrating Understanding & 0.47 & 0.57 & 0.61 & 0.39 & 0.13 & 0.43 & 0.42 \\
     & Advice Giving & 0.69 & 0.47 & 0.48 & 0.51 & 0.27 & 0.29 & 0.23 \\
     & Self-Oriented & 0.70 & 0.53 & 0.50 & 0.52 & 0.30 & 0.50 & 0.40 \\
     & Dismissing Emotions & 0.49 & 0.12 & 0.11 & 0.11 & 0.05 & 0.14 & 0.06 \\
     \bottomrule
    \end{tabular}%
    }

    \caption{\textbf{Expert-LLM Inter-rater Reliability across Models and Prompting Styles.} Inter-rater reliability between different LLMs and experts (median) for fewshot and Chain-of-Thought (CoT) prompts.}
    \label{ED:llms-kappa-median}
\end{table}

\newpage
\section{LLMs' Inter-Rater Reliability  Across Sub-components}\label{ED:llm-sec}

\begin{figure}[htbp]
    \centering
    \includegraphics[width=.9\linewidth]{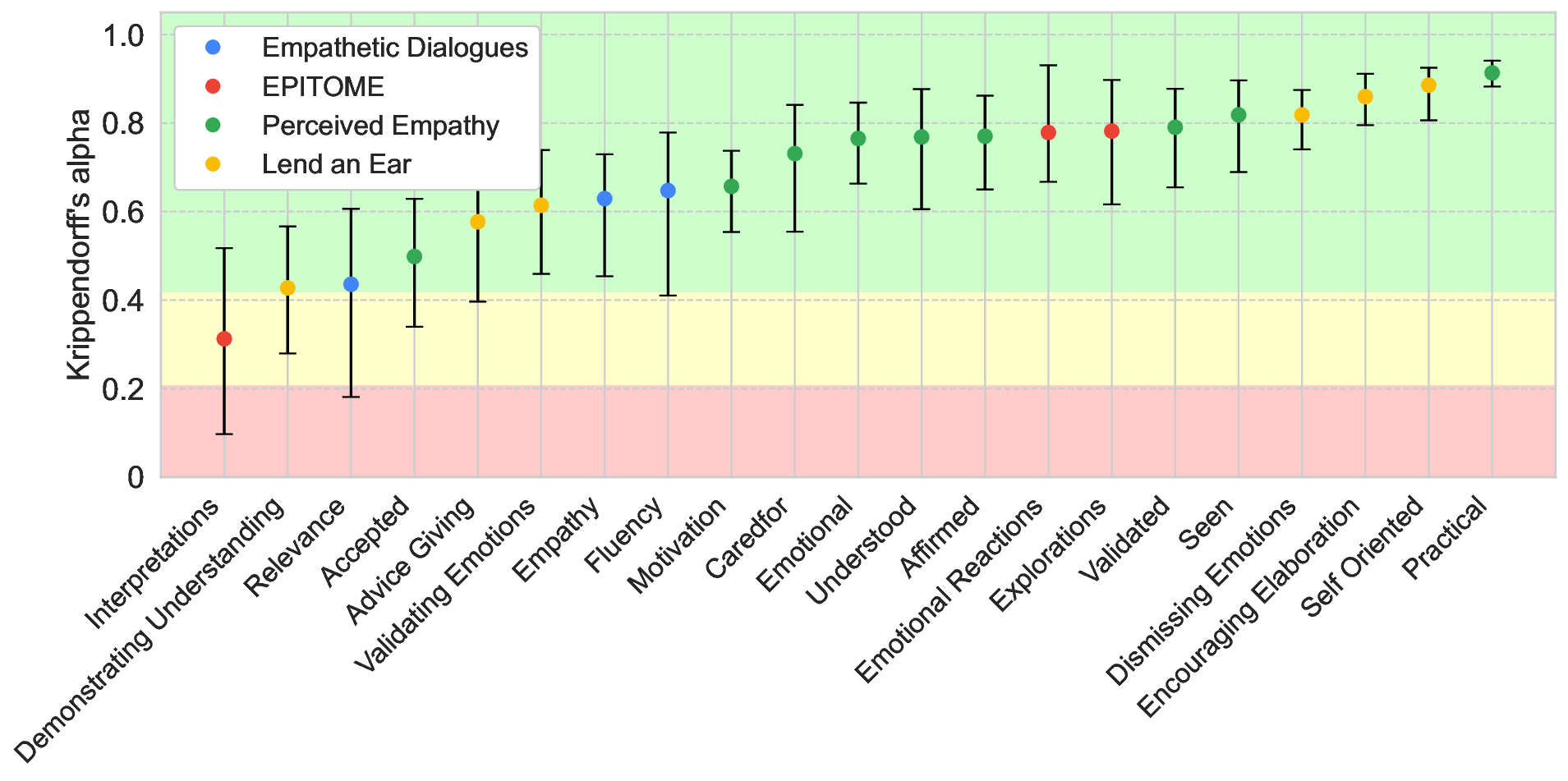}

    \caption{\textbf{LLM Inter-rater Reliability across Sub-components.} Inter-rater agreement as Krippendorff's alpha among the three LLM annotators (Gemini-2.5-pro-exp, GPT-4o, Claude3.7-sonnet)  across frameworks' sub-components. Error bars are 95\% confidence intervals obtained by bootstrapping conversations, and the background color indicates the degree of agreement}
    \label{fig:llm-kappa}
\end{figure}

\newpage
\section{Multicollinearity within Frameworks}\label{ED:VIF}

\begin{figure}[htbp]
    \centering
    \includegraphics[width=.9\linewidth]{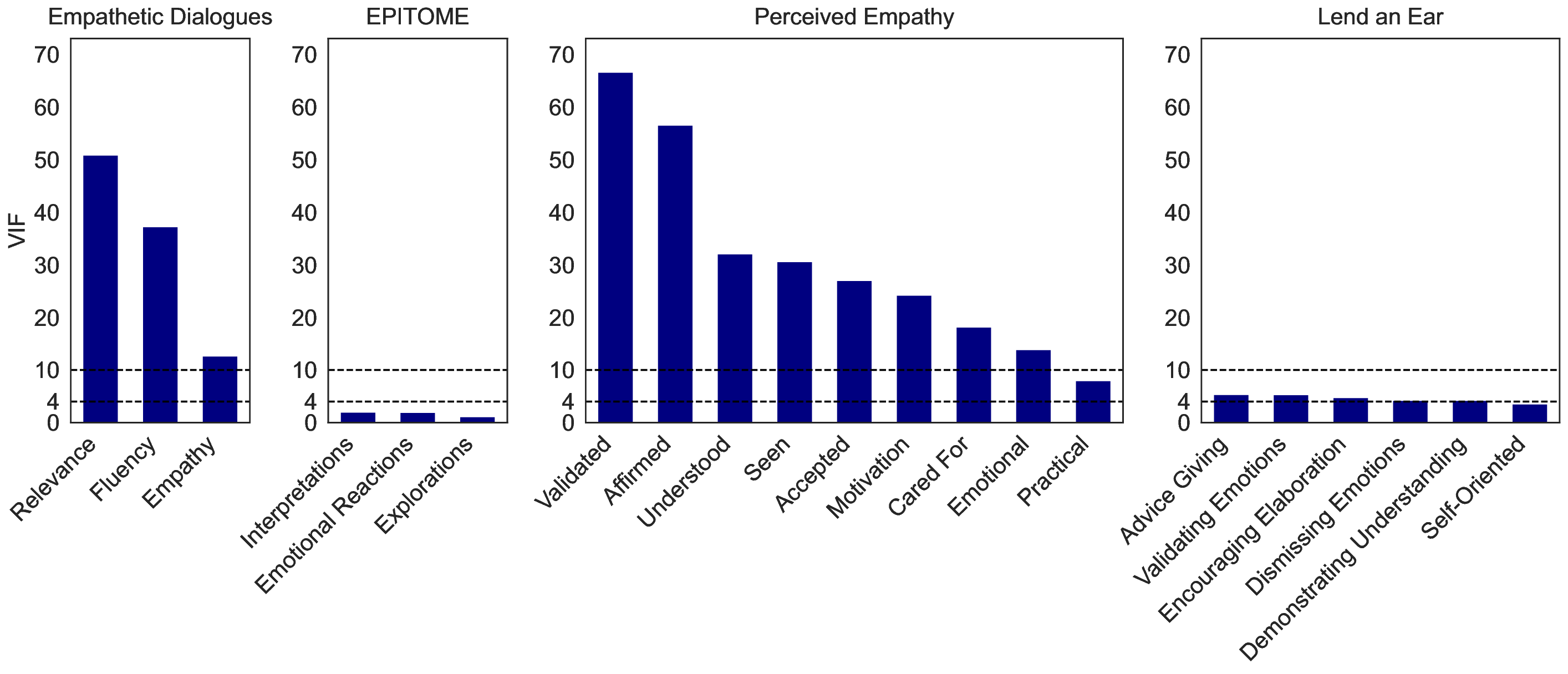}

    \caption{\textbf{VIF for each sub-component within the four frameworks.} VIF quantifies the degree of multicollinearity among sub-components, with higher values indicating greater redundancy. Dashed lines indicate conventional thresholds of VIF $\leq$ 4 indicating low multicollinearity, 4 $\leq$ VIF $\leq$ 10 indicating moderate multicollinearity, and VIF $\geq$ 10 indicating high multicollinearity \cite{regier2013dsm}. Sub-components with high VIF may indicate overlapping constructs or redundant measures within the framework.}
    \label{fig:vif}
\end{figure}

\newpage 
\section{Empathy Inflation by Crowds}\label{ED:crowds}

\begin{figure}[htbp]
    \centering
    \includegraphics[width=\linewidth]{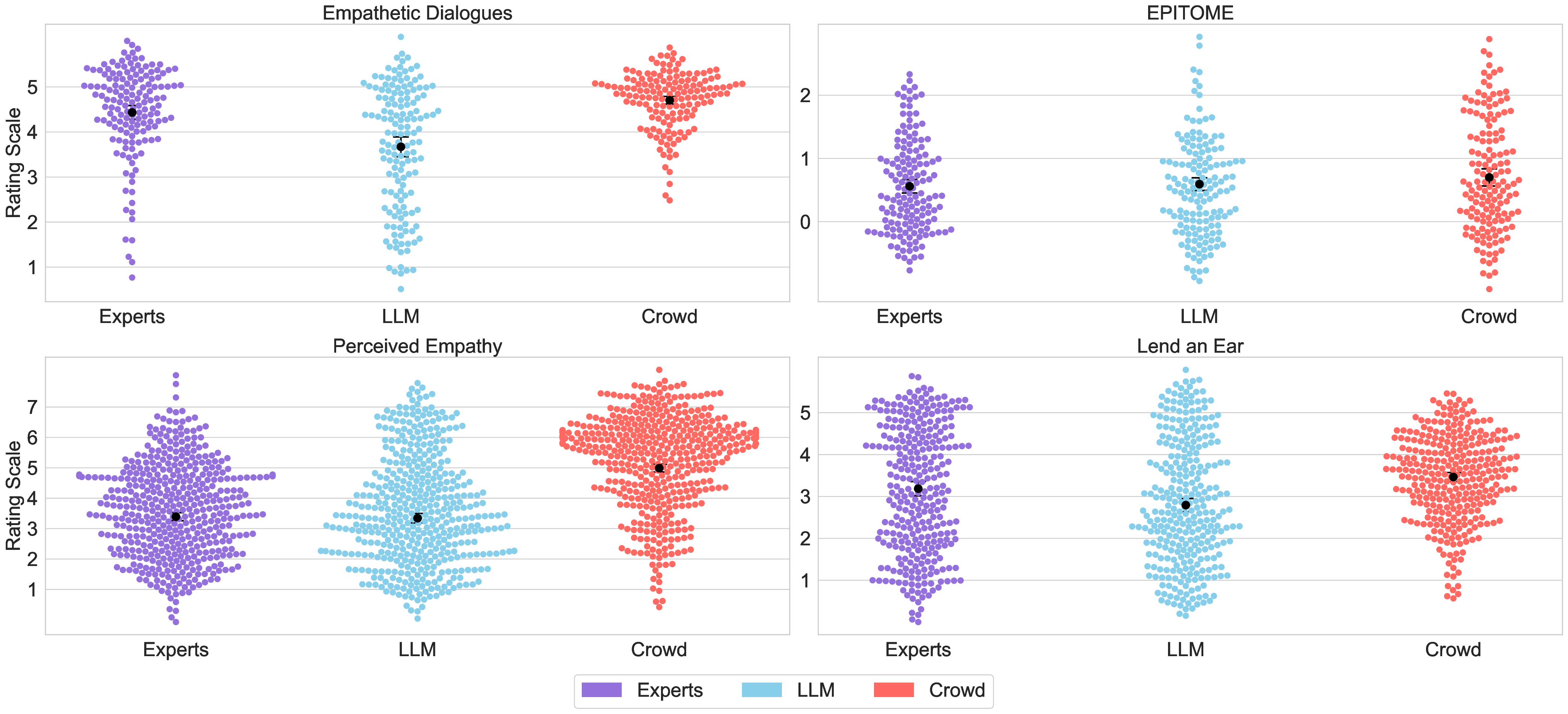}

    \caption{\textbf{Empathy Inflation by Crowds}. We present the distribution of annotations by experts, LLMs, and crowds across four evaluation frameworks on different scales. Empathetic Dialogues (1-5), EPITOME (0-2), Perceived Empathy (1-7), and Lend an Ear (1-5). Crowd workers tend to assign higher ratings than experts in all frameworks, with LLM ratings typically falling between the two. Dashed lines indicate mean ratings.}
    \label{fig:crowd-empathy-inflation}
\end{figure}

\newpage
\section{Self-reported Perceived Empathy to Crowd and Expert Evaluations}
\begin{figure}[htbp]
    \centering
    \includegraphics[width=0.7\linewidth]{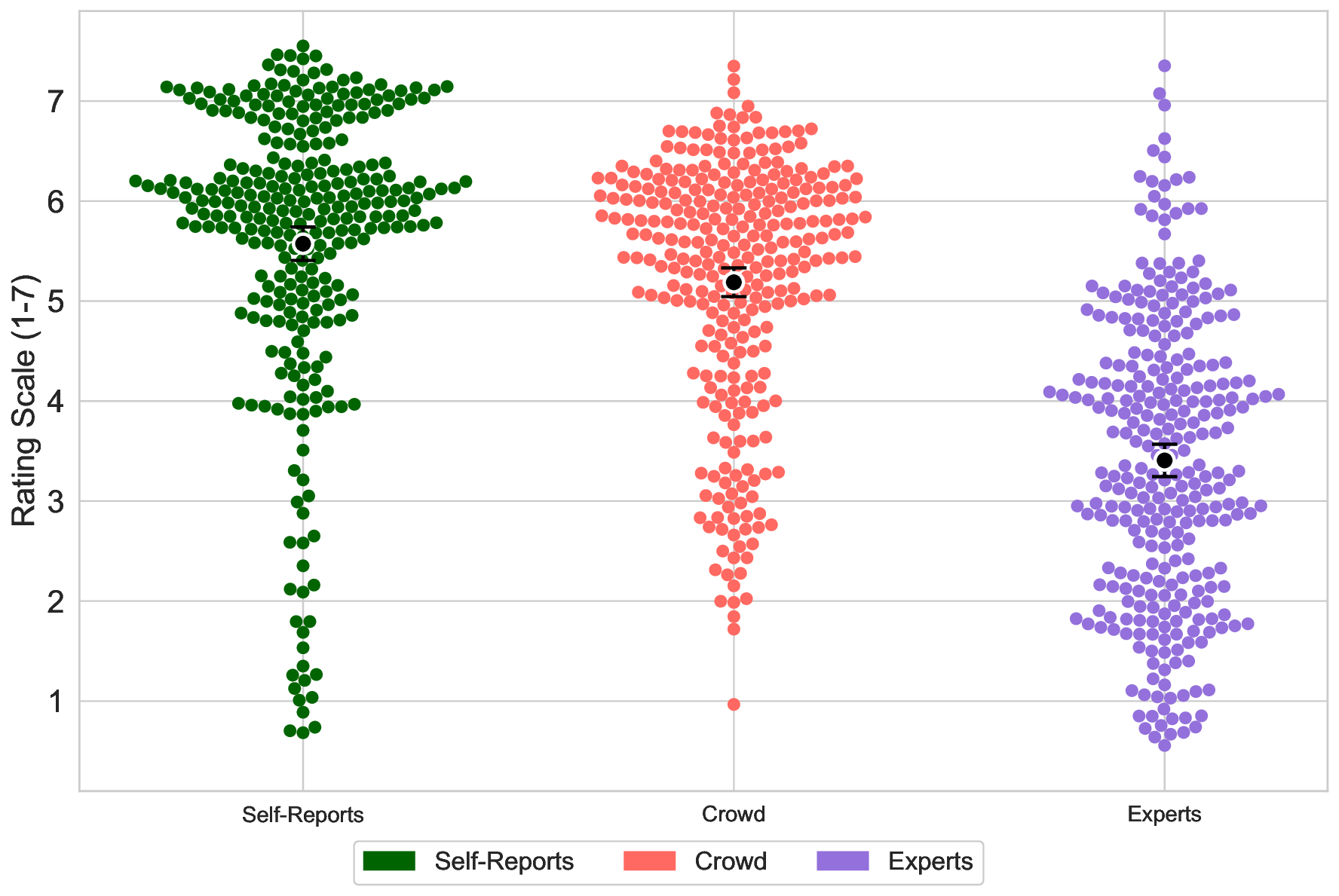}

    \caption{\textbf{Comparing Self-reported Perceived Empathy to Crowd and Expert Evaluations.} We present the distribution of self-reported perceived empathy and annotations from crowd and experts across six sub-components of the Perceived Empathy dataset.}
    \label{fig:selfvsothers}
\end{figure}

\section{LLM Prompt}\label{Supplemental Information:AIPrompt}
\begin{tcolorbox}[
  enhanced jigsaw,
  breakable,
  colback=green!5!white,
  colframe=green!50!black,
  title={\centering LLM Prompt},
  coltitle=black,
  fonttitle=\bfseries,
  titlerule=0mm,
  title style={top color=green!30!white, bottom color=green!10!white},
  coltitle=black,
  arc=3mm,
  boxrule=0.8pt,
  width=\textwidth,
  left=5pt,
  right=5pt,
  top=8pt,
  bottom=8pt,
  before skip=10pt,
  after skip=10pt
]
You are an empathic communication expert. You are assessing empathic support provided by the supporter to the seeker. Use the following framework to guide your assessment.            

\paragraph{Asking Open-Ended vs. Closed-Ended Questions}
\begin{itemize}
    \item Example
    \begin{itemize}[label=]
        \item Seeker: ''I’ve been feeling so detached from everything lately. Like, I’m just going through the motions without really being there.”
        \item Supporter: ''Do you feel like you’re going through depression?”
    \end{itemize}
    \item Annotation
    \begin{itemize}[label=--]
        \item \textit{What Happened:} The supporter shows some curiosity about the seeker’s state (''Are you depressed?”), but the question is closed-ended (yes/no). This tends to limit the seeker’s elaboration.
        \item \textit{More Empathic Alternative:}
        \begin{itemize}[label=--]
            \item Validate: ''That sounds really tough. It sounds like you feel numb or disconnected.”
            \item Open-Ended: ''Can you tell me more about what feeling detached is like for you?”
        \end{itemize}
        This approach acknowledges the seeker’s emotional distress and invites them to share more.
    \end{itemize}
\end{itemize}

\paragraph{Validating Before Problem-Solving}
\begin{itemize}
    \item Example
    \begin{itemize}[label=]
        \item Seeker: ''I feel overwhelmed and exhausted all the time, like I’m running on empty.”
        \item Supporter: ''What ways have you tried to recharge yourself?”
    \end{itemize}
    \item Annotation
    \begin{itemize}[label=--]
        \item \textit{What Happened:} The supporter immediately offers a problem-solving angle (''How are you recharging?”) rather than pausing to validate the seeker’s exhaustion.
        \item \textit{Why It’s Suboptimal:} It overlooks the intense emotions (''overwhelmed,” ''running on empty”) just mentioned.
        \item \textit{More Empathic Alternative:}
        \begin{itemize}[label=--]
            \item Mirror/Validate: ''Overwhelmed and running on empty—wow, that sounds really draining.”
            \item Empathic Follow-Up: ''What’s been weighing on you the most lately?”
        \end{itemize}
        By first reflecting the seeker’s emotional language and showing understanding, the supporter helps the seeker feel heard and accepted before moving on to any next step.
    \end{itemize}
\end{itemize}

\paragraph{Avoiding Unsolicited Advice}
\begin{itemize}
    \item Example
    \begin{itemize}[label=]
        \item Seeker: ''I used to do yoga, but I haven’t had the time or energy lately. I’m so drained.”
        \item Supporter: ''I think sometimes it’s important to push ourselves to enjoy things we used to—yoga could be so relaxing.”
    \end{itemize}
    \item Annotation
    \begin{itemize}[label=--]
        \item \textit{What Happened:} The supporter gives direct advice (''push ourselves” to do yoga) instead of focusing on how drained the seeker feels.
        \item \textit{Why It Feels Off:} Even if the advice is well-intentioned, the seeker just said they have no energy. This comes across as not fully listening.
        \item \textit{More Empathic Alternative:}
        \begin{itemize}[label=--]
            \item Validate: ''It must be discouraging when even yoga, something you used to love, feels like a chore.”
            \item Ask Permission to Advise: ''I have some ideas that might help—would you like to hear them, or do you just need me to listen right now?”
        \end{itemize}
        Soliciting permission respects the seeker’s autonomy and ensures they actually want advice.
    \end{itemize}
\end{itemize}

\paragraph{''It’s Not About the Nail”: Listening vs. Fixing}
\begin{itemize}
    \item Example
    \begin{itemize}[label=]
        \item Seeker: ''I’ve tried taking short breaks, but nothing seems to help much. I’m so overwhelmed.”
        \item Supporter: ''Have you reached out to a therapist? Maybe problem-solving with them would help.”
    \end{itemize}
    \item Annotation
    \begin{itemize}[label=--]
        \item \textit{What Happened:} The supporter jumps into ''fix it” mode (therapist referral), which can be helpful—if the seeker asked for solutions.
        \item \textit{Why It Can Miss the Mark:} The seeker just revealed deep distress (''overwhelmed”), which needs emotional acknowledgment.
        \item \textit{More Empathic Alternative:}
        \begin{itemize}[label=--]
            \item Mirror: ''Taking breaks hasn’t helped, and you’re still feeling overwhelmed—that must be so frustrating.”
            \item Explore: ''Can you say more about what’s weighing on you the most?”
        \end{itemize}
    \end{itemize}
\end{itemize}

\paragraph{Normalizing Struggle and Encouraging Without ''Shoulds”}
\begin{itemize}
    \item Example
    \begin{itemize}[label=]
        \item Seeker: ''Part of me feels like I should be okay by now. Maybe talking to someone could help.”
        \item Supporter: ''I think you should. You’ll come out with a new mindset.”
    \end{itemize}
    \item Annotation
    \begin{itemize}[label=--]
        \item \textit{What Happened:} The supporter is telling the seeker what they ''should” do (''go to therapy”), which can feel pushy.
        \item \textit{Why ''Should” Can Alienate:} It positions the supporter as an authority figure and can sound like ''I know better than you.”
        \item \textit{More Empathic Alternative:}
        \begin{itemize}[label=--]
            \item Validate and Normalize: ''It’s completely understandable to feel hesitant or think you ‘should’ be okay. We all need help sometimes.”
            \item Encourage Autonomy: ''If you’re open to it, talking with a therapist can be a safe place to figure out these overwhelming feelings.”
        \end{itemize}
    \end{itemize}
\end{itemize}

\paragraph{Expressing Concern and Care (Non-Abandonment)}
\begin{itemize}
    \item Example
    \begin{itemize}[label=]
        \item Seeker: ''I’m feeling alone. I don’t want to bother anyone with my problems.”
        \item Supporter: ''Is there anyone else you can talk to about this? Maybe a friend or family member?”
    \end{itemize}
    \item Annotation
    \begin{itemize}[label=--]
        \item \textit{What Happened:} The question might be well-meaning, but it sidesteps the fact that the seeker is coming to you now.
        \item \textit{Why It Might Feel Off:} The seeker just expressed feeling alone and not wanting to be a burden. Asking if they’ve tried someone else can unintentionally push them away.
        \item \textit{More Empathic Alternative:}
        \begin{itemize}[label=--]
            \item Direct Compassion: ''I’m so sorry you feel alone. I’m here to listen if you want to share more.”
            \item Invite Elaboration: ''Could you tell me more about what feeling alone looks like for you day to day?”
        \end{itemize}
    \end{itemize}
\end{itemize}

\paragraph{Encouraging Exploration of Feelings}
\begin{itemize}
    \item Example
    \begin{itemize}[label=]
        \item Seeker: ''I just lost my job today. It feels like a punch to the gut. I’m scared and overwhelmed, especially at 52.”
        \item Supporter: ''I’m sure 52 is still young. You can do anything you put your mind to. Have you looked at unemployment yet?”        
    \end{itemize}
    \item Annotation
    \begin{itemize}[label=--]
        \item \textit{What Happened:} The supporter rushes into reassurance (''52 is young!”) and problem-solving (unemployment benefits).
        \item Why It May Feel Minimizing: The seeker’s fear and shock (''punch to the gut”) get glossed over.
        \item \textit{More Empathic Alternative:}
        \begin{itemize}[label=--]
            \item Acknowledge Shock: ''Oh no—losing a job like that must feel like such a blow. I’m so sorry you’re going through this.”
            \item Invite Reflection: ''What scares you most about this next chapter?” By acknowledging the impact first, you show you’re truly hearing the emotional weight of the situation
        \end{itemize}

    \end{itemize}
\end{itemize}

\paragraph{Key Takeaways}
\begin{itemize}[label=--]
    \item Validate First, Solve Later
    \item Before offering suggestions, reflect the emotion (''That sounds really tough”) so they feel heard.
    \item Ask Permission for Advice: ''Would you like some ideas, or do you just need me to listen right now?” This keeps the seeker’s agency intact.
    \item Use Open-Ended Questions. Instead of ''Are you depressed?” try ''What does feeling detached look like for you day to day?”
    \item Mirror Their Words: Repeat keywords or phrases (''overwhelmed,” ''running on empty,” ''punch to the gut”) to show you’ve genuinely absorbed what they said.
    \item Offer Empathy, Not Just Reassurance: Saying ''Oh, that’s scary. I’m here for you” goes further than ''You’ll be fine!” because it acknowledges real pain.
    \item Express Non-Abandonment: Phrases like ''I’m not going anywhere; I care about you” remind them they have support, even if no immediate solution exists.
\end{itemize}

Final Note: All of these examples underscore one major theme: when someone is hurting, the main goal is to make them feel seen and heard, not to instantly fix the problem. If you do give advice, ask if they want it first. Above all, empathic listening requires genuine curiosity about their experience, consistent emotional validation, and reassurance that you will remain present with them in their distress.

\textcolor{blue}{\textbf{Few Shot Examples}: Here we insert three example conversations and provide numeric expert annotations for the specific sub-component under evaluation.}

Now assess this conversation: \textcolor{blue}{\textbf{Conversation}: Here we insert a conversation from the dataset}

\textcolor{blue}{\textbf{Question and Grading Rubric}: Here we insert the exact question and grading rubric for the specific sub-component under evaluation.}

Respond with ONLY the number.
\end{tcolorbox}

\newpage
\section{Illustrative examples reveal when experts and LLMs diverge from crowdworkers}\label{sec:qual-analysis}

\begin{figure}
    \centering
    \includegraphics[width=0.85\linewidth]{SupplementaryFigure1.eps}
    \caption{\textbf{Example conversations and annotations.} Radar plots visualizing expert, LLM, and crowd annotations for example conversations from (A) the Lend an Ear dataset (1 = not at all, 5 = very much; VE = Validating Emotions, DE = Dismissing Emotions, SO = Self-Oriented, AG = Advice Giving, DU = Demonstrating Understanding, EE = Encouraging Elaboration) and (B) the EPITOME dataset (0 =No Communication, 1 = Weak Communication, and 2 = Strong Communication).}
    \label{fig:example_convos}
\end{figure}

Experts' and crowdworkers' annotations and their accompanying explanations highlight systematic differences in how sub-components within each framework were interpreted.\footnote{This qualitative analysis (see Methods for more details) of explanations is based on the Lend an Ear and EPITOME datasets because those are the only two datasets that include explanations alongside the annotations.} For responses identified by median experts as dismissive  in the Lend an Ear dataset, at least one crowd annotator interpreted the response as either validating emotions or demonstrating understanding in 48\% of the conversations. In contrast, for these conversations, no pair of expert annotators ever disagreed in this way, i.e, one expert never interpreted a response as dismissive while another interpreted it as validating or demonstrating understanding. In EPITOME, crowds identified strong communication of interpretations in 50\% of conversations, compared to only 20\% by experts. These discrepancies suggest that crowd annotators lacked nuance or insufficiently considered conversational context when evaluating conversations. 

In order to qualitatively reveal differences between annotator types, we present two examples where experts' and LLMs' annotations differ significantly from the crowd's. Figure \ref{fig:example_convos} presents an example from the Lend an Ear dataset. Experts and LLMs annotate this conversation as lacking a single instance of ``Encouraging Elaboration'' or ``Demonstrating Understanding''. In contrast, the crowd reported higher scores on these sub-components with ratings of 2 (Slightly) and 3 (Somewhat) on 5-point Likert scales, respectively. Experts identified \emph{Yea, we all wish things were different...} as ``Dismissing Emotions'' because it de-individualizes the seeker’s negative experience by framing it as something everyone experiences. In contrast, crowd annotations varied where one marked this response as ``Validating Emotions'', another as ``Encouraging Elaboration'', a third as ``Demonstrating Understanding'', and a fourth as ``Dismissing Emotions''. 

Figure \ref{fig:example_convos}B offers a second example from the EPITOME dataset that reveals how experts and LLMs are aligned on annotations but the crowdworker annotations differ. Experts and crowds marked the response \emph{Anyone in your life you could reach out to?} as an instance of ``Explorations'', but they differ on the degree. Experts all considered this an instance of ``Weak Exploration'' noting its superficiality and the way it pushed the effort of support back onto the seeker or their support network. In contrast, the crowd annotator annotated this as an instance of ``Strong Exploration'', which is the maximum score possible. Similarly, experts and crowds identified the response as an instance of ``Interpretations'' (communicating an understanding of the seeker's experiences and feelings). However, experts consider this a weak instance whereas the crowdworker considers it a strong instance. The experts identify the phrase ``\emph{Being alone is difficult}'' as justification for this annotation because this phrase paraphrases the situation and sentiment expressed by the support seeker. In contrast, the crowd annotator identified ``\emph{I reconnected with an old friend}'' as justification despite this phrase not being a direct communication around feelings. See Figures~\ref{fig:egconv1}, \ref{fig:egconv2}, and \ref{fig:egconv3} in the Supplemental Information for additional example conversations and annotations that further illustrate discrepancies between expert, crowd, and LLM annotations.

\clearpage
\section{Example Conversations and Annotations}\label{Supplemental Information:examples}

\begin{figure}[h]
    \centering
    \includegraphics[width=\linewidth]{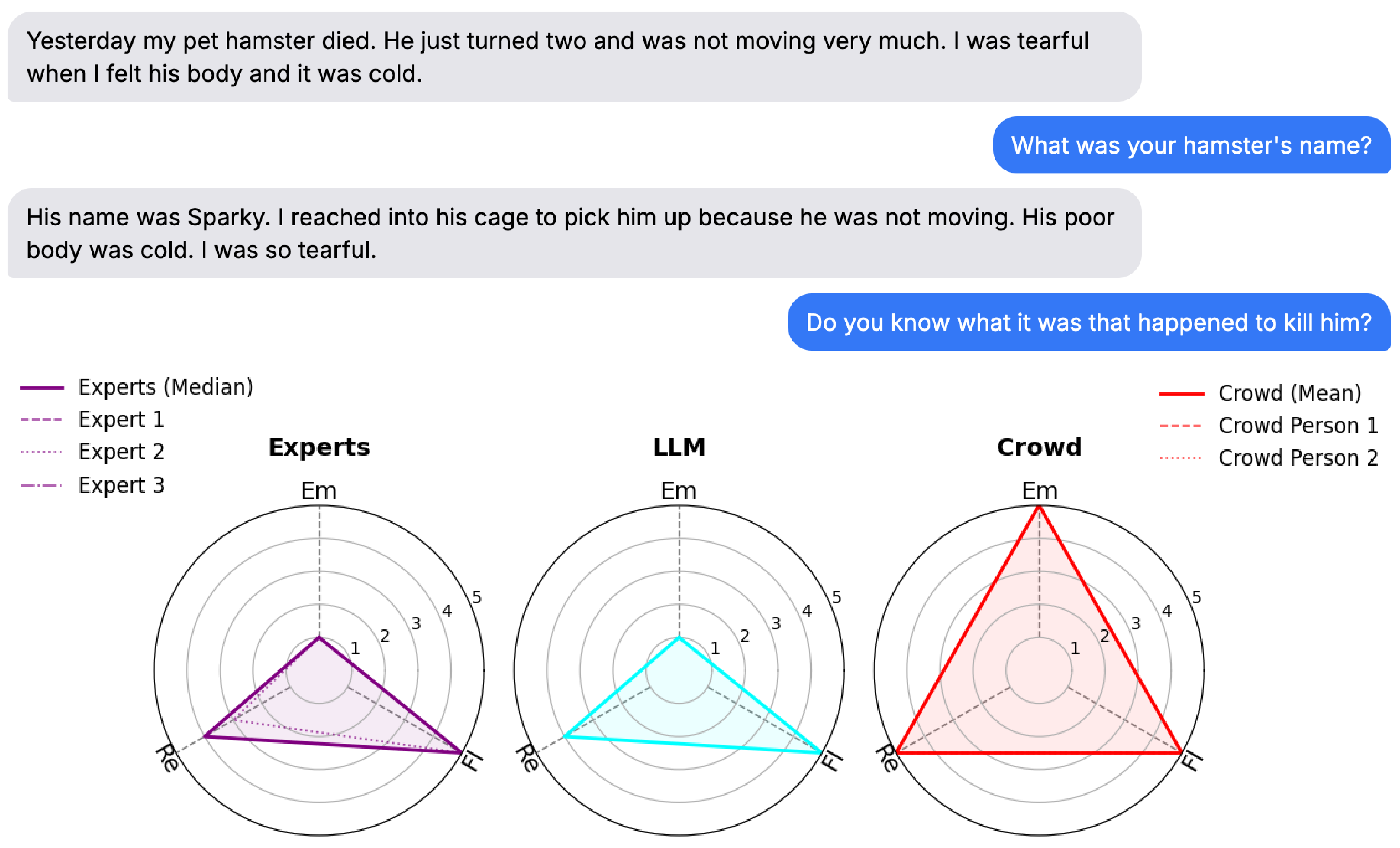}
    \caption{\textbf{Example Conversation and Annotations from the Empathetic Dialogues Dataset.} Radar plots visualizing expert, LLM, and crowd annotations for an example conversation from the Empathetic Dialogues dataset (1 = not at all, 5 = very much; Em = Empathy, Fl = Fluency'', Re = Relevance).}
    \label{fig:egconv1}
\end{figure}
\begin{figure}[h]
    \centering
    \includegraphics[width=\linewidth]{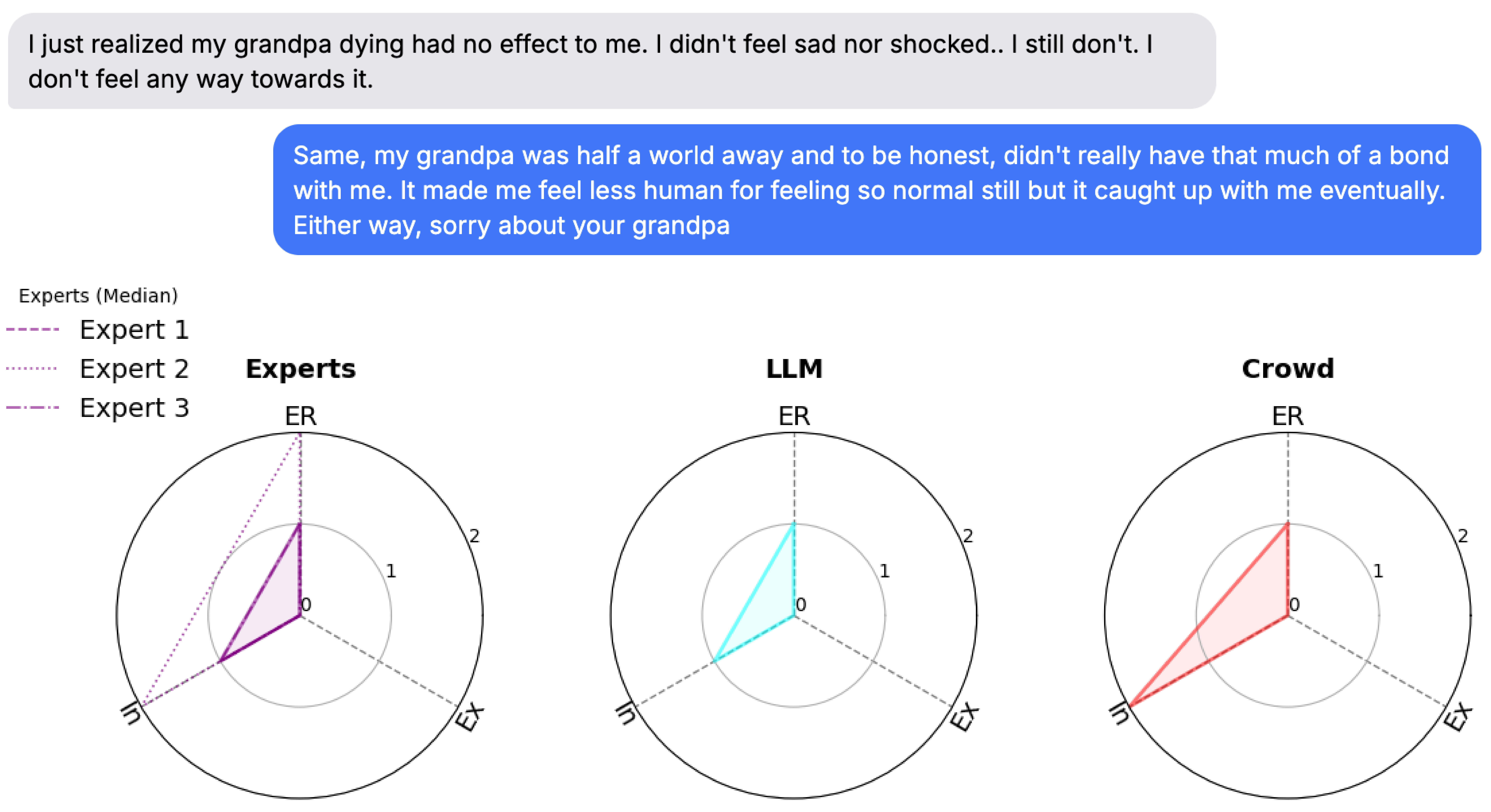}
    \caption{\textbf{Example Conversation and Annotations from the EPITOME Dataset.} Radar plots visualizing expert, LLM, and crowd annotations for an example conversation from the EPITOME dataset 0 = No Communication, 1 = Weak Communication, and 2 = Strong Communication; ER = Emotional Reactions, Ex = Explorations, In = Interpretations).}
    \label{fig:egconv2}
\end{figure}
\begin{figure}[h]
    \centering
    \includegraphics[width=\linewidth]{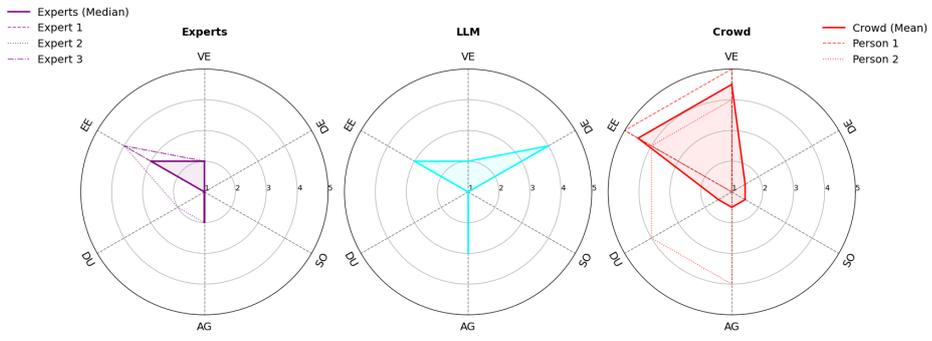}
    \caption{\textbf{Example Conversation and Annotations from the Lend an Ear Dataset.} Radar plots visualizing expert, LLM, and crowd annotations for an example conversation from the Lend an Ear dataset (1 = not at all, 5 = very much; VE = Validating Emotions, DE = Dismissing Emotions, SO = Self-Oriented, AG = Advice Giving, DU = Demonstrating Understanding, EE = Encouraging Elaboration).}
    \label{fig:egconv3}
\end{figure}

\clearpage

\section{Robustness of LLM Annotations}

\subsection{Variability across Runs}
We evaluate run-rerun reliability of LLM annotations~\cite{barrie2024replication} by re-annotating each dataset five times with temperature fixed at 0. We find that the correlation of annotations between runs is $0.93$, which indicates that the LLM annotations are not fully deterministic but highly deterministic in aggregate.

\subsection{Inconsistency across Contexts}
We evaluate inconsistency across contexts of LLM annotations~\cite{gu2024survey} by examining differences in annotations across 50 conversations in the three role-playing scenarios including feeling overlooked and undervalued, job loss, or being passed over for a promotion (see section on Role Playing Scenarios for exact scenario prompts) in the Lend an Ear pilot dataset. We conduct Kruskal-Wallis tests for each of six dimensions, which allows us to evaluate whether the distribution of LLM annotations for each sub-component differs across the three standardized role‑playing contexts. We find no evidence of inconsistencies in LLM annotations across conversational contexts in the Lend an Ear pilot dataset. Table \ref{tab:inconsistencies} details the Kruskal–Wallis statistics and Benjamini--Hochberg FDR-adjusted p-values for each dimension.

\begin{table}[ht]
\centering
\begin{tabular}{l c c}
\hline
\textbf{Sub-component} & \textbf{Kruskal-Wallis H} & \textbf{p value (adjusted)} \\
\hline
Validating emotions         & 2.706 & 0.431 \\
Encouraging elaboration     & 6.798 & 0.200 \\
Demonstrating understanding & 2.496 & 0.431 \\
Advice giving               & 1.241 & 0.562 \\
Self oriented               & 1.152 & 0.562 \\
Dismissing emotions         & 3.106 & 0.431 \\
\hline
\end{tabular}
\caption{\textbf{Testing for inconsistency in LLM annotations across three role-playing contexts.} Kruskal-Wallis test statistic with Benjamini--Hochberg FDR-adjusted p-values of six annotation sub-components across three Lend an Ear scenarios.}
\label{tab:inconsistencies}
\end{table}

\subsection{Miscalibrated Confidence} 
LLMs provide only numeric scores rather than free-form justifications or confidence judgments, which avoids concerns about miscalibrated confidence~\cite{Steyvers2025}. 

\subsection{Verbosity Bias}
We evaluate verbosity bias  defined as the tendency for humans and LLMs to favor longer responses in their annotations~\cite{ye2024justice, Steyvers2025}, by examining how the number of words shared by a supporter correlates with expert and LLM ratings. Table \ref{tab:verbosity_effects} presents ordinary least squares (OLS) estimates of verbosity effects for 21 sub-components of empathic communication. The Experts column shows how verbosity (word count) influences expert annotations, and the LLM column shows how verbosity influences LLM annotations.  

We find evidence of verbosity bias in both human experts and LLMs with verbosity bias even more prevalent in human experts than LLMs. Experts showed statistically significant (at the $p<0.05$ level with p-values adjusted via the Benjamini-Hochberg procedure) verbosity bias on 14 of the 21 sub-components, compared to only 3 for LLMs. In terms of effect size of verbosity bias, we find an increase of 10 words led to a 0.2 to 0.3 point increase in expert ratings of whether the seeker was understood, validated, affirmed, seen, accepted, or cared for. We also find two negative correlations: experts' annotations of fluency and LLMs' annotations of dismissing emotions.

\begin{table}[ht]
\centering
\begin{tabular}{llcc}
\toprule
\textbf{Framework} & \textbf{Sub-component} & \textbf{Experts} & \textbf{LLM} \\
\midrule
\multirow{3}{*}{Empathetic Dialogues} & Empathy & 0.038 (0.000) & 0.038 (0.015) \\
 & Relevance & 0.011 (0.027) & 0.014 (0.592) \\
 & Fluency & -0.010 (0.001) & -0.005 (0.773) \\
\midrule
\multirow{3}{*}{EPITOME} & Emotional Reactions & 0.003 (0.000) & 0.001 (0.592) \\
 & Interpretations & 0.003 (0.000) & 0.002 (0.587) \\
 & Explorations & -0.000 (0.863) & -0.001 (0.773) \\
\midrule
\multirow{9}{*}{Perceived Empathy} & Understood & 0.018 (0.000) & -0.002 (0.824) \\
 & Validated & 0.021 (0.000) & -0.004 (0.773) \\
 & Affirmed & 0.021 (0.000) & -0.003 (0.814) \\
 & Seen & 0.021 (0.000) & -0.001 (0.895) \\
 & Accepted & 0.025 (0.000) & -0.004 (0.773) \\
 & Cared For & 0.024 (0.000) & 0.005 (0.768) \\
 & Emotional & 0.014 (0.000) & 0.016 (0.084) \\
 & Practical & 0.031 (0.000) & 0.021 (0.083) \\
 & Motivation & 0.034 (0.000) & 0.029 (0.000) \\
\midrule
\multirow{6}{*}{Lend an Ear} & Validating Emotions & 0.004 (0.022) & 0.009 (0.015) \\
 & Encouraging Elaboration & 0.001 (0.496) & 0.000 (0.948) \\
 & Demonstrating Understanding & 0.001 (0.410) & 0.004 (0.079) \\
 & Advice Giving & 0.002 (0.152) & 0.003 (0.618) \\
 & Self-Oriented & 0.004 (0.001) & 0.007 (0.083) \\
 & Dismissing Emotions & 0.002 (0.066) & -0.011 (0.024) \\
\bottomrule
\end{tabular}
\caption{\textbf{Verbosity Bias in Expert and LLM Annotations with FDR control.} OLS regression coefficients with Benjamini--Hochberg FDR-adjusted p-values in parentheses ($N = 50$ observations).}
\label{tab:verbosity_effects}
\end{table}

\subsection{Apparently sensible yet wrong answers}
We evaluate apparently sensible yet wrong LLM annotations~\cite{zhou2024larger} by looking at cases in which LLM annotations diverge from expert annotations. Specifically, we examine conversations where median expert rating and median LLM rating differ by at least one point on a three point Likert Scale or three or more points on a five or seven point Likert scale. 

Such discrepancies were observed across all datasets. For instance, in the Empathetic Dialogues dataset (Figure~\ref{empdiag_aswa}), we found one case where experts and LLMs diverged by three points on the fluency sub-component. In this example, the seeker expressed disappointment about a football game (\emph{``I am not happy with how the Eagles played yesterday''}), and while the supporter initially stayed on topic, they soon shifted to an irrelevant remark (\emph{``I can't believe I made this much money with bitcoin''}). Experts judged the supporter’s message as quite fluent (score of 4), but LLMs rated it as minimally fluent (score of 1).

In the Lend an Ear pilot dataset (Figure~\ref{fig:rpg_aswa}), we identified five conversations with discrepancies of three or more points. In one case, a seeker described feeling discouraged after being passed over for a promotion. Experts noted that the supporter dismissed the seeker’s emotions with remarks such as \emph{``be strong for yourself''}, assigning a rating of 2 (somewhat) on the dismissing emotions sub-component. In contrast, the LLM annotation gave the same response a rating of 5 (very much), substantially overstating the degree of dismissal.  

In the EPITOME dataset, while 26 out of 50 conversations had a one-point difference on the three-point scale between expert and llm ratings, there were no cases with a two-point difference. Note that even a one-point gap is important in this context as it reflects a shift from reporting the complete absence of a behavior (score of 0) to identifying a weak presence of it (score of 1). For example, in Figure~\ref{fig:epitome_aswa}, a seeker expressed suicidal ideation (\emph{``Currently sitting at the water's edge, wanting to jump. Don't worry I don't think I actually will.''}). The supporter responded with clarifying questions (\emph{``Why? Do you want to talk about it? What’s made you feel like wanting to?''}). Both experts and LLMs labeled the response as highly exploratory (score of 2). However, the divergence emerged on emotional reactions and interpretations: experts judged the response as entirely lacking these components (score of 0), while LLMs annotated a weak presence of both (score of 1).  

We also identified five conversations in the Perceived Empathy dataset where expert and LLM ratings differed by three or more points. However, since the conversational data for this dataset is only available upon request~\cite{yin2024ai}, we do not present detailed examples here.

These examples highlight that while LLM annotations often appear reasonable, they can deviate from expert judgments in both magnitude and interpretation.

\begin{figure}[htbp]
    \centering
    \includegraphics[width=\linewidth]{SupplementaryFigure5.eps}
    \caption{Radar plots visualizing expert, LLM, and crowd annotations for example conversations from the Empathetic Dialogues dataset (1 = not at all, 5 = very much; Em = Empathy, Re = Relevance, Fl = Fluency)}
    \label{empdiag_aswa}
\end{figure}

\begin{figure}[htbp]
    \centering
    \includegraphics[width=\linewidth]{SupplementaryFigure6.eps}
    \caption{Radar plots visualizing expert, LLM, and crowd annotations for example conversations from the Lend an Ear dataset (1 = not at all, 5 = very much; VE = Validating Emotions, DE = Dismissing Emotions, SO = Self-Oriented, AG = Advice Giving, DU = Demonstrating Understanding, EE = Encouraging Elaboration)}
    \label{fig:rpg_aswa}
\end{figure}

\begin{figure}[htbp]
    \centering
    \includegraphics[width=\linewidth]{SupplementaryFigure7.eps}
    \caption{Radar plots visualizing expert, LLM, and crowd annotations for example conversations from the EPITOME dataset; (0 = No Communication, 1 = Weak Communication, and 2 = Strong Communication; ER = Emotional Reactions, In = Interpretations, Ex = Explorations)}
    \label{fig:epitome_aswa}
\end{figure}

\newpage

\clearpage
\section{Role playing scenarios in the Lend an Ear pilot experiment}\label{role-playing-scenarios}
\begin{tcolorbox}[
  enhanced jigsaw,
  breakable,
  colback=blue!5!white,
  colframe=blue!50!black,
  title={``So, I just lost my job today. I had a sense this was coming, but it's still a shock..''},
  coltitle=black,
  fonttitle=\bfseries,
  titlerule=0mm,
  title style={top color=blue!30!white, bottom color=blue!10!white},
  arc=3mm,
  boxrule=0.8pt,
  width=\textwidth,
  left=5pt,
  right=5pt,
  top=8pt,
  bottom=8pt,
  before skip=10pt,
  after skip=10pt
]
Emily is a 52-year-old who was an HR manager at a health company until she just got laid off. Emily had a feeling something was up—there were rumors about layoffs, and she noticed the usual signs like leadership changes and budget cuts. But when she got the email saying her job was being cut, it still felt like the rug was pulled out from under her. The Zoom call with her boss that followed was awkward, with him barely looking her in the eye.
Even though she kind of saw it coming, actually losing her job hit her hard. She just sat there, staring at her computer, feeling like all those years of work disappeared in an instant. 
The idea of dusting off her resume after so many years felt overwhelming, and starting over at 52? That is downright scary.
Emily feels stuck between feeling relieved that the stress of what will happen next is over (and she doesn't have to deal with her annoying manager) but being terrified about what comes next. Would anyone even want to hire someone her age? And how long would it take to find something new? The thought of becoming irrelevant in her field was creeping in.
Friends tried to cheer her up with comments like, “You’ve got tons of experience; you’ll find something in no time!” or “Maybe now you can finally take a break.” But what she really needed was for someone to just get it, to say something along the lines of “I know this is tough. It’s okay to feel scared and unsure about what’s next.”
\end{tcolorbox}

\begin{tcolorbox}[
  enhanced jigsaw,
  breakable,
  colback=blue!5!white,
  colframe=blue!50!black,
  title={``I've been feeling so detached from everything lately, like I'm just going through the motions without really being there.''},
  coltitle=black,
  fonttitle=\bfseries,
  titlerule=0mm,
  title style={top color=blue!30!white, bottom color=blue!10!white},
  arc=3mm,
  boxrule=0.8pt,
  width=\textwidth,
  left=5pt,
  right=5pt,
  top=8pt,
  bottom=8pt,
  before skip=10pt,
  after skip=10pt
]
Alex is a 33-year-old software developer at Google who has recently been promoted to Senior Software Engineer with a significant pay raise and increased visibility.
Despite the pay raise and increased visibility, she’s feeling the weight of burnout. She’s constantly exhausted, surviving on energy drinks and power naps. Her once-weekly yoga sessions have dwindled to none, and she hasn’t had a full night’s sleep in months. Conversations with her spouse have become strained, mostly revolving around logistics rather than connection. She feels an increasing sense of guilt every time she hears her children ask, “When will you be home, Mommy?”
The family photo on her desk, once a source of comfort, now feels like a reminder of the growing distance between her and her loved ones.
She’s torn between the fear of losing her hard-earned status at Google and the fear of losing the connection with her family. The thought of stepping back or saying no feels impossible, yet the idea of continuing at this pace seems equally unbearable.
She’s frustrated by well-meaning but surface-level advice like "just take a break" or "it’ll get better soon." What she craves is a real conversation, someone who can see beyond her professional façade and understand the quiet desperation she feels. 
She’s yearning for someone to say something along the lines of “I see how hard this is for you,” rather than offering quick solutions.
\end{tcolorbox}

\begin{tcolorbox}[
  enhanced jigsaw,
  breakable,
  colback=blue!5!white,
  colframe=blue!50!black,
  title={``I'm feeling so discouraged.. I just got passed over for the promotion I was working hard for''},
  coltitle=black,
  fonttitle=\bfseries,
  titlerule=0mm,
  title style={top color=blue!30!white, bottom color=blue!10!white},
  arc=3mm,
  boxrule=0.8pt,
  width=\textwidth,
  left=5pt,
  right=5pt,
  top=8pt,
  bottom=8pt,
  before skip=10pt,
  after skip=10pt
]
Marcus, a 25-year-old Black man working in a marketing position at a mid-sized company. Marcus had been working tirelessly in his marketing role, staying late to perfect campaigns, volunteering for challenging projects, and even coming up with innovative ideas that gained positive attention from clients. He had his eye on a promotion to Product Marketing Manager, a role he felt he had more than earned. But when the promotion was announced, Marcus was crushed to find out it went to someone else—someone who hadn’t put in the same hours or shown the same dedication.
He feels so discouraged. It’s hard not to think that maybe his hard work isn’t being noticed because of something he can’t control, like his race. He was so close to getting what he’s been working for, and now it just feels like a huge setback. It’s not just about the promotion—it’s about believing that hard work actually pays off.
The job he used to be excited about now feels like a grind.
Marcus is torn between pushing himself even harder to prove his worth or just coasting through the day. He’s starting to wonder if all the effort is really worth it or if he’s just being overlooked because of who he is. The whole situation leaves him feeling stuck and unsure about what to do next.
When he vents to his friends, they say things like, “Don’t worry, your time will come,” or “Maybe it wasn’t the right fit.” But Marcus doesn’t need that right now. He just wants someone to get how frustrating this is and say something along the lines of “It’s okay to be upset—you worked hard, and it sucks when that’s not recognized.”

\end{tcolorbox}

\clearpage
\section{Crowd annotator instructions in the Lend an Ear pilot experiment}

\begin{figure}
    \centering
    \includegraphics[width=0.8\linewidth]{SupplementaryFigure8.eps}
    \caption{Instructions given to crowd annotators before annotating conversations from the Lend an Ear dataset.}
    \label{fig:instructions}
\end{figure}

\clearpage
\section{Comparing Perceived Empathy with Judgments of Empathy Communication}\label{Supplemental Information:selfvsother}

We compared support seekers' self-reported perceptions of empathy with annotator evaluations (crowdworkers and experts) across six of the nine sub-components (``understood", ``validated", ``affirmed", ``seen", ``accepted", and ``cared for") in the Perceived Empathy dataset, where self-report data were available. Figure \ref{fig:selfvsothers} illustrates the distributions of self-report, crowd, and median expert ratings. We observe `empathy inflation' in participants' self-reports as they  predominantly use the upper end of the rating scale (scores of 5, 6, and 7). 

We conducted a non-parametric correlation analysis using Kendall's $\tau_b$ which accounts for ties and evaluates whether groups rank responses in the same order. Across the 50 conversations, we found no meaningful differences between self-crowd and self-expert alignment. Kendall's $\tau_b$ was 0.16 (95\% CI [$0.06$, $0.26$]) for self--crowd and 0.16 (95\% CI [$0.06$, $0.25$]) for self--expert comparisons, based on bootstrap confidence intervals. At the dimension level, the pattern was mixed: crowd ratings were closer to self-reports for ``understood", whereas expert ratings were closer for ``validated" and ``cared for". Median correlations were nearly identical between crowds (median $\tau_b = 0.14$) and experts (median $\tau_b = 0.14$).

These results indicate that while crowd annotations appear superficially similar to self-reports, this is largely driven by both groups inflating their ratings toward the high end of the Likert scale. This pattern is consistent with social desirability bias, as both support-seekers and crowdworkers tend not to use the full range of the scale. Once ordinal agreement is taken into account, we find no evidence that crowdworkers systematically track participants’ own assessments better than trained experts.

\end{appendices}
\end{document}